\documentclass[10pt,twocolumn,letterpaper]{article}

\usepackage{cvpr}              %
\usepackage{multirow}
\usepackage{multicol}
\usepackage{color, colortbl}
\usepackage{soul}
\usepackage{comment}
\usepackage{amsmath}
\usepackage{amssymb}
\usepackage{tikz}
\usepackage{subcaption} %

\usepackage{pifont}%

\usepackage{pgfplots}
\usepackage{color}
\usetikzlibrary{spy}

\definecolor{cvprblue}{rgb}{0.21,0.49,0.74}
\definecolor{Gray}{gray}{0.9}
\definecolor{darkgreen}{RGB}{5, 120, 15}
\definecolor{lightgreen}{HTML}{9CCBB8}
\definecolor{lightred}{HTML}{E3242B}
\definecolor{lightorange}{HTML}{ED7D31}

\usepackage[pagebackref,breaklinks,colorlinks,allcolors=cvprblue]{hyperref}
\usepackage{caption}
\def\confName{ICCV}
\def\confYear{2025}

\title{Evaluating Variance in Visual Question Answering Benchmarks}

\author{Nikitha SR\\
Media and Data Science Research Lab, Adobe\\
{\tt\small nikithasr@adobe.com}}

\begin{document}

\maketitle

\setlength{\textfloatsep}{1pt}
\begin{abstract}
Multimodal large language models (MLLMs) have emerged as powerful tools for visual question answering (VQA), enabling reasoning and contextual understanding across visual and textual modalities. Despite their advancements, the evaluation of MLLMs on VQA benchmarks often relies on point estimates, overlooking the significant variance in performance caused by factors such as stochastic model outputs, training seed sensitivity, and hyperparameter configurations. This paper critically examines these issues by analyzing variance across 14 widely used VQA benchmarks, covering diverse tasks such as visual reasoning, text understanding, and commonsense reasoning. We systematically study the impact of training seed, framework non-determinism, model scale, and extended instruction finetuning on performance variability. Additionally, we explore Cloze-style evaluation as an alternate assessment strategy, studying its effectiveness in reducing stochasticity and improving reliability across benchmarks. Our findings highlight the limitations of current evaluation practices and advocate for variance-aware methodologies to foster more robust and reliable development of MLLMs. 

\end{abstract}
    
\section{Introduction}
\label{sec:intro}

Vision language models are a cornerstone of artificial intelligence bridging textual and visual modalities by different mechanisms and mimicking human intelligence. This multimodal capability enables them to analyse images and videos in the context of text, handle open ended queries over visually grounded tasks. Multimodal large language models are a subclass of these models that are built by incorporating visual processing ability into existing large language models. These models combine the language model abilities like reasoning along with visual processing to perform a wide range of tasks like image captioning, visual reasoning and visual question answering. A huge body of existing literature \cite{liu2024improvedbaselinesvisualinstruction, chen2024internvlscalingvisionfoundation, li2024llavaonevisioneasyvisualtask, lin2024vilapretrainingvisuallanguage, liu2025nvilaefficientfrontiervisual, tong2024cambrian1fullyopenvisioncentric,ye2024mplugowl3longimagesequenceunderstanding,lu2024deepseekvlrealworldvisionlanguageunderstanding, bai2023qwenvlversatilevisionlanguagemodel, deitke2024molmopixmoopenweights, hong2024cogvlm2visuallanguagemodels, yao2024minicpmvgpt4vlevelmllm,marafioti2025smolvlmredefiningsmallefficient} has proposed ways to combine the visual information (usually through an encoder) with the language model that generates text. A generic structure of these models involve a vision encoder that provides encoded visual tokens which are projected into the LLM's textual token space through different projection mechanisms.

Multimodal large language models are thus sophisticated AI systems with comprehensive world knowledge combining the powers of dual modality. These models are used in conversation style interfaces primarily for visual question answering (VQA), where the goal is to answer natural language questions based on visual content such as images or videos. MLLMs can interpret complex scenes, understand context, and generate coherent, informative responses. In VQA tasks, MLLMs are particularly effective at handling diverse question types, ranging from object recognition and counting to more advanced reasoning involving spatial relationships, actions, or even commonsense understanding. Their ability to align visual and textual modalities enables them to provide more accurate and context-aware answers, making them valuable for applications in accessibility, education, healthcare, and robotics. As MLLMs continue to improve, they push the boundaries of what machines can perceive and reason about, making VQA a key area of focus in developing general-purpose, multimodal AI systems. Subsequently, a huge plethora of VQA benchmarks \cite{singh2019vqamodelsread, hudson2019gqanewdatasetrealworld, goyal2017makingvvqamatter} exist to assess MLLM performance addressing different aspects like fine grained visual understanding\cite{wu2023vguidedvisualsearch, mathew2021docvqadatasetvqadocument}, factual or descriptive questions, counting, binary yes/no queries\cite{li2023evaluatingobjecthallucinationlarge}, spatial reasoning\cite{hudson2019gqanewdatasetrealworld}, and activity recognition. More complex types involve commonsense reasoning, compositional understanding, and knowledge-based questions\cite{ying2024mmtbenchcomprehensivemultimodalbenchmark} that require linking visual cues with external world knowledge.

Evaluation benchmarks are central to assessing the empirical efficacy of machine learning methods. However, a persistent and often overlooked issue is the presence of variance in benchmark results, which can lead to misleading conclusions particularly when improvements are incremental. This concern is especially relevant in the context of Visual Question Answering (VQA) benchmarks, where performance is influenced by multiple sources of variability, including the stochastic nature of large language model (LLM) generations and sensitivity to hyperparameters.

Despite these factors, the majority of multimodal large language models (MLLMs) report only point estimates on VQA benchmarks, without accounting for performance variability. In this paper, we critically examine this practice by analyzing the variance and its contributing factors across several widely used VQA benchmarks. Our goal is to highlight the limitations of point-estimate-based evaluation and to advocate for more rigorous, variance-aware assessment practices, thereby fostering a more grounded and reliable foundation for MLLM development.

\section{Experimental Setup}

To perform our analysis, we use the training framework of LLaVA-1.5 and InternVL2.5. These frameworks employ a modular architecture that integrates a vision encoder into an LLM via a lightweight MLP projector.  
\begin{figure}
    \centering
    \includegraphics[width=1\linewidth]{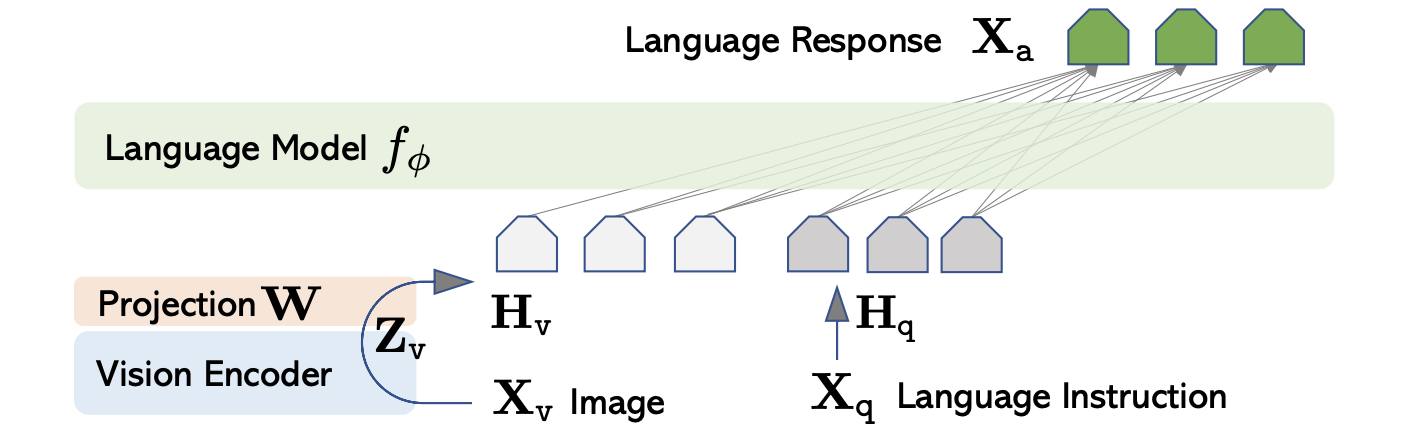}
    \caption{LLaVA1.5 architecture}
    \label{fig:llava-arch}
\end{figure}
\begin{figure}
    \centering
    \includegraphics[width=1\linewidth]{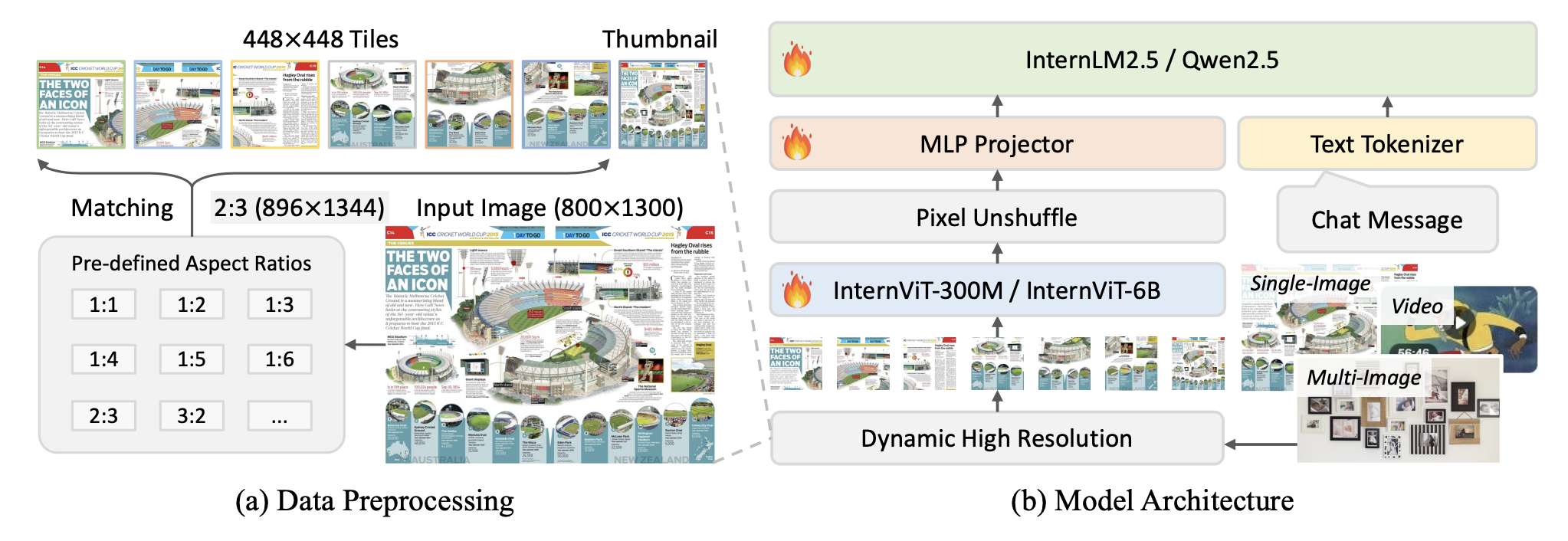}
    \caption{InternVL2.5 architecture}
    \label{fig:internvl-arch}
\end{figure}
Formally, we construct the Vision Language Model by combining the language model $f_{\theta}$ with vision encoder $Z_v$ through the lightweight projection layer $W$. The image is processed through $Z_v$ and the resulting visual features are projected by $W$ into the LLM’s embedding space. These projected visual tokens are then concatenated with the text tokens $H_q$ to form the model input. While both the frameworks constitute similar overall network architecture (ViT-Projector-LLM)\ref{fig:internvl-arch}\ref{fig:llava-arch}, they differ in the training paradigms, training data and image processing techniques. Broadly their training paradigms constitute 2 stages,
\begin{itemize}
    \item  \textbf{Pretraining}: This phase aims to align the vision tokens with text embedding space of the LLM. This is done by training only the projector $W$ while keeping the rest of the MLLM frozen. Image-captioning data is used to perform the alignment. The data is converted into instruction following format as proposed in \cite{liu2024improvedbaselinesvisualinstruction}. InternVL2.5 \cite{chen2024internvlscalingvisionfoundation} also includes question answering datasets for its pretraining stage.
    \item \textbf{Instruction-finetuning}: In this phase, either the entire model or select components (typically the projector and parts of the LLM) are finetuned to follow instruction-style prompts, particularly for VQA tasks. Since the LLM is unfrozen during this stage, careful curation of the finetuning data is essential to prevent undesirable or unstable model behavior at inference time.
\end{itemize}

To evaluate the generalizability of the observed variance, we conduct our analysis by fully training LLaVA-1.5 and performing stage-2 finetuning of InternVL2.5.  We do not perform full training for InternVL2.5, as the complete training datasets used in its pretraining stage is not publicly available. In the following sections, we demonstrate that even modifying a single stage of the training pipeline can lead to significant performance variance, highlighting the sensitivity of MLLMs to changes in training procedures.

Unless otherwise noted, all variance measurements are computed over 5 independent runs with identical hyperparameters and different random seeds. For non-deterministic training, each configuration was repeated 5 times using the same seed to capture stochastic variance due to framework-level non-determinism.
\subsection{Benchmarks}

We evaluate our models across a comprehensive suite of 14 widely adopted visual question answering (VQA) benchmarks, each designed to assess distinct vision-language capabilities. VQAv2\cite{goyal2017makingvvqamatter} targets language bias and visual grounding by providing complementary image-question pairs, while GQA\cite{hudson2019gqanewdatasetrealworld} focuses on compositional reasoning using scene graphs from Visual Genome. SEED-Bench\cite{li2023seedbenchbenchmarkingmultimodalllms} offers a multi-choice format to test a wide range of VQA skills, including spatial reasoning, counting, and OCR. To evaluate real-world robustness, we include VizWiz\cite{gurari2018vizwizgrandchallengeanswering}, sourced from questions asked by blind users, and Pope\cite{li2023evaluatingobjecthallucinationlarge}, which probes object hallucinations through binary object presence questions. For text and document-based understanding, we use TextVQA\cite{singh2019vqamodelsread} to assess scene text reading, DocVQA\cite{mathew2021docvqadatasetvqadocument} for document structure comprehension, and InfographicsVQA\cite{mathew2021infographicvqa} for reasoning over visually rich infographic layouts. ChartQA\cite{masry2022chartqabenchmarkquestionanswering} and AI2D\cite{Kembhavi2016ADI} further test logical reasoning on charts and diagrams, respectively. To evaluate academic and scientific knowledge, we include ScienceQA\cite{lu2022learn}, a multi-choice dataset covering natural and social sciences, and MMMU\cite{yue2024mmmumassivemultidisciplinemultimodal}, which contains college-level, multi-disciplinary multimodal questions. Finally, MM-Vet\cite{yu2024mmvetevaluatinglargemultimodal} serves as a small but challenging GPT-graded benchmark combining multiple reasoning types, and V$^*$\cite{wu2023vguidedvisualsearch} focuses on fine-grained object attribute and spatial relation understanding in high-resolution scenes. Together, these benchmarks span a diverse range of tasks, formats, and evaluation protocols, providing a robust testbed for assessing multimodal model capabilities.

\section{Experiments}
We systematically study the performance variance across the above diverse suite of 14 benchmarks. We consider various factors that could influence the sensitivity of the MLLMs like the training seed, training non-determinism, LLM size, and continued instruction finetuning. We also analyse if Cloze-style evaluation leads to reduction in variance as suggested in \cite{madaan2024quantifyingvarianceevaluationbenchmarks} by evaluating on MMMU-Cloze, ScienceQA-Cloze, SEED-Bench-Cloze, TextVQA-Cloze, Pope-Cloze and V$^*$-Cloze.

\subsection{Role of Seed}
For a benchmark, with a preferred metric  $\mathcal{S}$, and a set of models $\mathbb{M} ={\mathbb{M}_1, \mathbb{M}_2, . . . \mathbb{M}_N}$, we define the benchmark variance $\sigma(\mathcal{S},\mathbb{M})$ as the standard deviation of the
metric $\mathcal{S}$ scores ${\mathcal{S}_M = \mathcal{S}_{M1}, \mathcal{S}_{M2}. . . \mathcal{S}_{Mn} }$ for each of the models in $\mathbb{M}$.
We report the performance across 5 different trainings by varying the training seed while keeping the remaining setup fixed. We report the test set size, performance by chance, mean accuracy $\mu(\mathcal{S},\mathbb{M})$, and performance variance $\sigma(\mathcal{S},\mathbb{M})$ in \ref{tab:main-var}. Except for DocVQA and InfographicsVQA, where variance is reported in terms of ANLS, and for POPE, where F1-score variance is used, all other benchmarks report accuracy-based variance. Note that VQAv2 and VizWiz are evaluated on their respective test-dev splits.

\subsubsection{Observations}
We observe a notable performance improvement across benchmarks on InternVL2.5. However, InfoVQA, MM-Vet, and V$^*$ show relatively low scores across models, with V$^*$ near chance for LLaVA-1.5. LLaVA-1.5 also exhibits high variance on MM-Vet (1.08\%), VizWiz (1.38\%), and V$^*$ (4.22\%), possibly due to a lack of robustness or over-sensitivity to certain image/question types and also the small dataset sizes. Most other benchmarks have a consistent 0.2-0.4\% variance with VQAv2 showing the least variance (0.04\%). Comparing with InternVL2.5 on the same scale (ie, 8B model) we observe increased variance across several benchmarks like ScienceQA ($0.28\rightarrow0.61$), TextVQA ($0.25\rightarrow0.74$) and MM-Vet ($1.08\rightarrow3.09$). Remaining benchmarks show variance in the similar range of $0.2$ to $0.8$. 

\begin{table*}[t]
\centering
\begin{tabular}{l|r|r|cc|cc|cc|cc|cc}
\toprule
\textbf{Benchmark} & \textbf{Size} & \textbf{Chance} & \multicolumn{2}{c}{\textbf{LLaVA-7B}} & \multicolumn{2}{c}{\textbf{InternVL-1B}} & \multicolumn{2}{c}{\textbf{InternVL-2B}} & \multicolumn{2}{c}{\textbf{InternVL-4B}} & \multicolumn{2}{c}{\textbf{InternVL-8B}}\\
\cmidrule(lr){4-13}
 & & & $\mu$ & $\sigma$ & $\mu$ & $\sigma$ & $\mu$ & $\sigma$ & $\mu$ & $\sigma$ & $\mu$ & $\sigma$ \\
\hline
GQA & 12578 & 0 & 62.62 & 0.20&-&-&-&-&-&-&-&- \\
ScienceQA & 4241 & 37.5 & 69.3 & 0.28 &82.53&0.51&82.79&1.44&91.78&0.88&91.71&0.61 \\
TextVQA & 5000 & 0 & 58.34 & 0.25&59.51&0.43&62.41&0.32&65.9&0.21&65.26&0.74 \\
Pope-Rand & 2910 & 50 & 87.14 & 0.27&85.58&0.35&87.36&0.47&86.16&0.23&87.26&0.15 \\
Pope-Pop & 3000 & 50 & 85.6 & 0.2&84.26&0.3&85.92&0.33&84.94&0.18&85.84&0.21 \\
Pope-Adv & 3000 & 50 & 84.24 &0.37&83.24&0.23&84.84&0.43&84.12&0.19&84.52&0.15  \\
SEED-Bench & 14233 & 25 & 66.41 & 0.32&68.68&0.32& 71.06&0.33&73.12&0.24&74.43&0.3\\
MM-Vet & 217 & 0 & 31.7 & 1.08 &32.42&0.63&31.74&0.8&40.56&1.22&39.94&3.09\\
DocVQA & 5186 & 0 & 21.47 & 0.49&52.73&0.22&55.13&0.3&57.46&0.25&56.39&0.33 \\
InfoVQA & 3288 & 0 & 21.07 & 1.14&22.98&0.2&24.11&0.37&31.1&0.67&31.12&0.43 \\
MMMU & 10495 & 25 & 31.7 & 0.22&-&-&-&-&-&-&-&-\\ 
VizWiz & 4000 & 0 & 44.8 & 1.38&38.52&0.63&40.42&0.82&50.59&0.75&50.47&0.83 \\
V$^*$-Attr & 115 & 25 & 43.65 & 2.25 &-&-&-&-&-&-&-&-\\
V$^*$-Pos & 76 & 50 & 54.47 & 4.22&-&-&-&-&-&-&-&- \\
VQAv2 & - & 0 & 79.16 & 0.04&48.93&0.06&49.77&0.07&50.73&0.06&50.59& 0.07\\
ChartQA & 2500 & 0 & - & -&74.98&1.08&78.7&0.95&79.07&0.37&55.68&0.47 \\
AI2D & 3088 & 25 & - & -&67.14&0.59&71.86&0.88&78.64&0.26&78.37&0.63 \\
\hline
\end{tabular}
\caption{Benchmark performance of LLaVA-1.5 and InternVL2.5 with mean and variance. Chance indicates the possible performance by chance.}
\label{tab:main-var}
\end{table*}

\subsection{Role of Non-deterministic Training}

\begin{figure}
    \centering
    \includegraphics[width=0.9\linewidth]{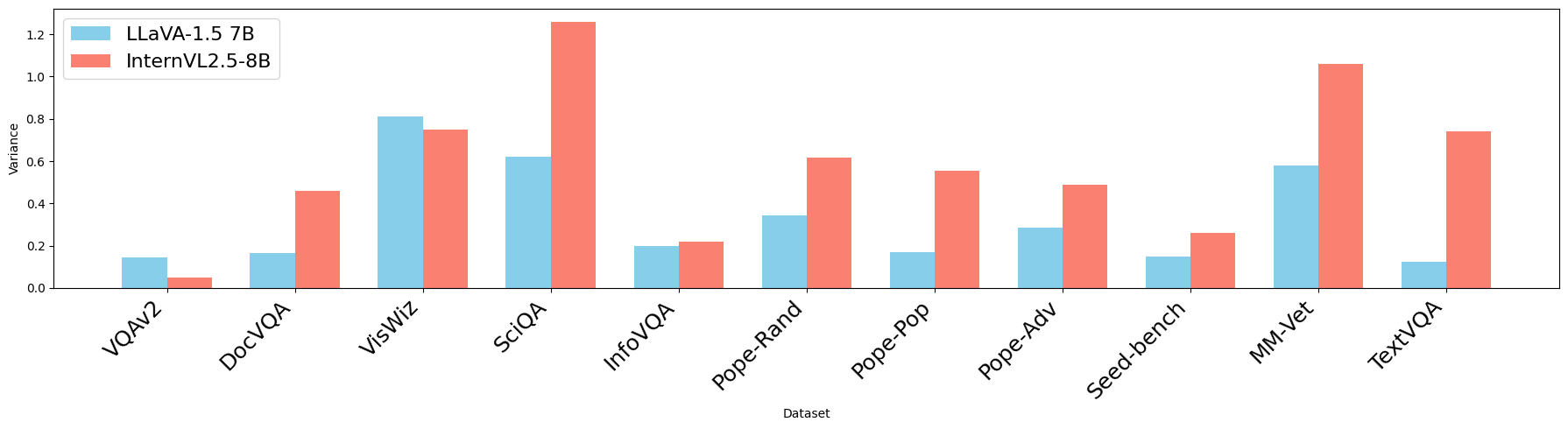}
    \caption{Comparison of variance between LLaVA-1.5-7B and InternVL2.5-8B trained with same seed. Increased variance across benchmarks for InternVL2.5-8B}
    \label{fig:frame_variance_same_seed}
\end{figure}

\begin{table*}[t]
\centering
\begin{tabular}{l|cc|cc|cc|cc|cc}
\toprule
\textbf{Benchmark} & \multicolumn{2}{c}{\textbf{LLaVA-7B}} & \multicolumn{2}{c}{\textbf{InternVL-1B}}& \multicolumn{2}{c}{\textbf{InternVL-2B}}& \multicolumn{2}{c}{\textbf{InternVL-4B}}& \multicolumn{2}{c}{\textbf{InternVL-8B}}\\
\cmidrule(lr){2-11}
 & $\mu$ & $\sigma$ & $\mu$ & $\sigma$ & $\mu$ & $\sigma$ & $\mu$ & $\sigma$ & $\mu$ & $\sigma$\\
\midrule

GQA & 61.26 & 0.07&-&-&-&-&-&-&-&- \\
ScienceQA  & 68.36 & 0.62 & 82.59 & 0.64 & 83.39 & 1.88 & 91.54 & 0.74 & 90.63 & 1.25\\
TextVQA  & 55.31 & 0.12&59.95&0.38&62.11&0.21&65.97&0.23&65.27&0.74\\
Pope-Rand  &  84.55 &0.34 & 85.64 & 0.3 & 87.22 & 0.24 & 86.14 & 0.44 & 87.6 & 0.62\\
Pope-Pop  & 83.32 & 0.29 &84.5&0.27&85.78&0.16&84.94&0.36&86.32&0.55\\
Pope-Adv  &  82.52 & 0.29&83.56&0.25&84.66&0.28& 84.14&0.32&84.78&0.48\\
SEED-Bench & 66.63 & 0.15 & 68.65 & 0.24 & 70.98 & 0.18 & 73.33 & 0.25 & 74.44 & 0.26\\
MM-Vet  & 29.65 & 0.58 &33.14&1.65&32&1.25&41.86&1.2&40.68&1.06\\
DocVQA & 21.22 & 0.16 & 52.53 & 0.33 & 55.17 & 0.26 & 57.43 & 0.23 & 56.3 & 0.46\\
InfoVQA & 19.78 & 0.2 & 23.07 & 0.36 & 24.32 & 0.19 & 30.46 & 0.4 & 31.2 & 0.22\\
MMMU  & 32.1 & 0.34&-&-&-&-&-&-&-&- \\ 
VizWiz  & 44.62 & 0.81 & 38.31 & 0.45 & 41.1 & 0.68 & 50.02 & 0.82 & 49.88 & 0.75\\
V$^*$-Attr  & 45.21 & 0.7&-&-&-&-&-&-&-&- \\
V$^*$-Pos  & 59.86 & 0.75&-&-&-&-&-&-&-&- \\
VQAv2   & 78.29 & 0.14 & 49.02 & 0.02 & 49.85 & 0.07 & 50.74 & 0.07 & 50.6 & 0.05\\
ChartQA & - & -&74.75&0.88&78.4&0.93&78.98&0.66&79.71&0.83 \\
AI2D & - & - &67.34&0.72&71.46&0.62&78.57&0.31&78.55&0.53\\
\midrule

\end{tabular}
\caption{Benchmark performance of LLaVA-1.5 and InternVL2.5 trained with the same seed.}
\label{tab:nondeterminism}
\end{table*}
One significant issue we faced while training MLLMs is the non-deterministic nature of the models. While the non-determinism would be expected to have little to nil effect on the model performance, we observe significant variance across benchmarks while run on the same seed. Hence we summarise the statistics associated to non-determinism by training LLaVA-1.5 and finetuning InternVL2.5 with a constant seed in \ref{tab:nondeterminism}. Often in literature \cite{chen2024recoverablecompressionmultimodalvision, shi2024needlargervisionmodels, sr2025hirelightweighthighresolutionimage}, we observed increments of 0.5-1\% accuracy to be attributed to various architectural modifications. However we see that certain benchmarks have variance large enough that could lead to false attributions. 
Benchmarks like ScienceQA and MM-Vet have variance higher than 1 even when the trainings are repeated with the same seed. High variance is also observed in ChartQA, VizWiz and AI2D. Comparing between frameworks on the same scale, we consistently find InternVL2.5 having higher variance than LLaVA-1.5. This is counterintuitive, as InternVL2.5 incorporates architectural stability and an enhanced instruction tuning pipeline. Yet, as shown in \ref{fig:frame_variance_same_seed}, variance across benchmarks increases despite only fine-tuning, suggesting potential trade-offs between generalizability and task-specific stability. In extreme cases, we observe performance jumps of up to 5\% in DocVQA purely due to non-determinism.

These observations prompted us to investigate the possible causes of such high variance. Specifically, we explore the following questions: i) Is the model capacity a factor in sensitivity?, ii) Are the evaluations formats (in terms of prompts) flawed? iii) Do certain characteristics of the benchmarks themselves contribute to this instability. 
\begin{figure}
    \centering
    \includegraphics[width=0.9\linewidth]{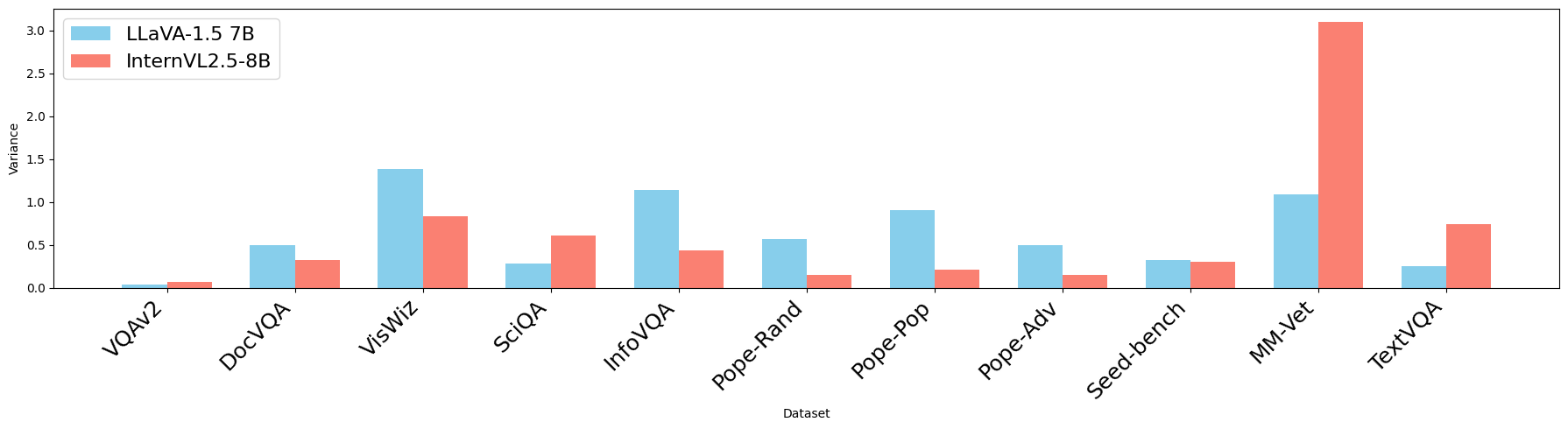}
    \caption{Comparison of variance between LLaVA-1.5-7B and InternVL2.5-8B trained with different seeds.}
    \label{fig:frame_variance}
\end{figure}
\subsection{Effect of Model Size}
\begin{figure}[htbp]
    \centering
    \begin{subfigure}[b]{1\linewidth}
        \includegraphics[width=\linewidth]{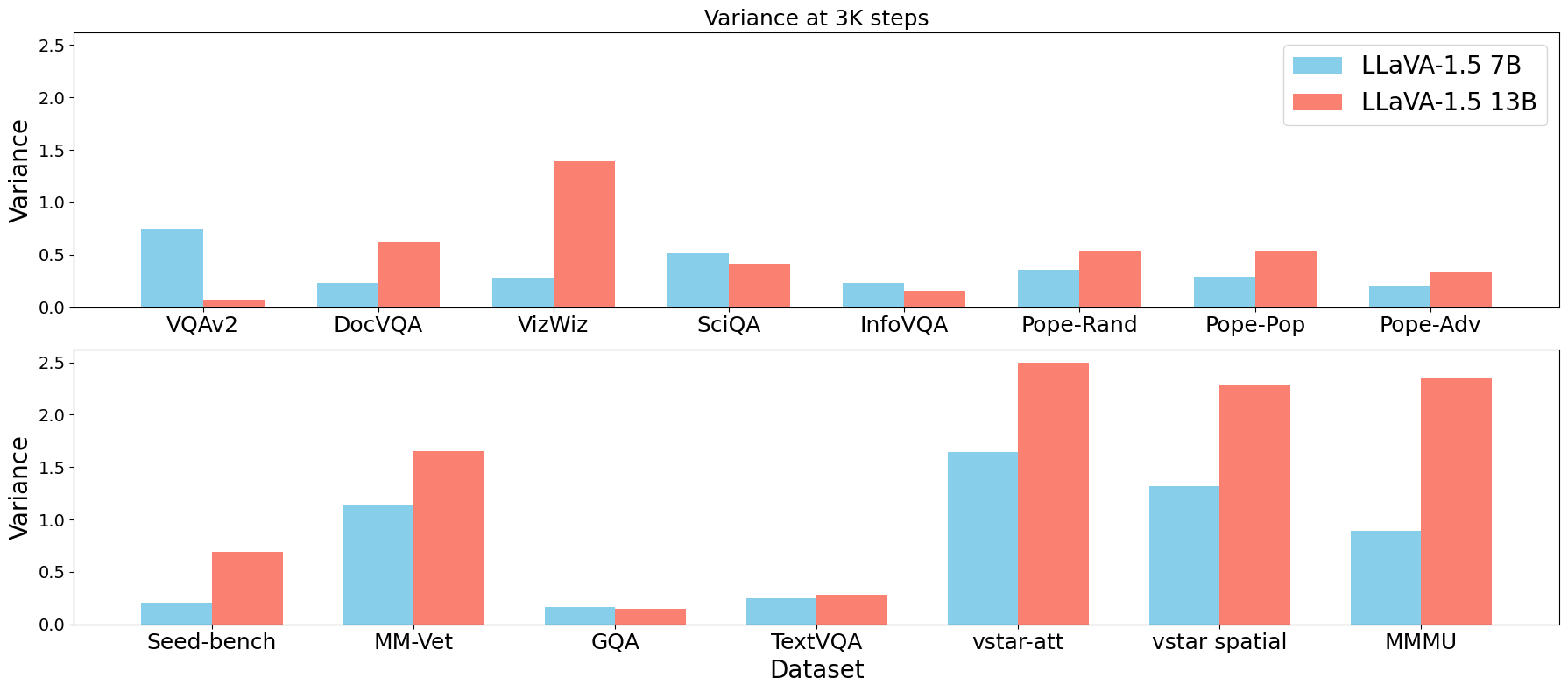}
        \caption{Variance comparison at 3K steps of finetuning.}
        \label{fig:sub1}
    \end{subfigure}
    \begin{subfigure}[b]{1\linewidth}
        \includegraphics[width=\linewidth]{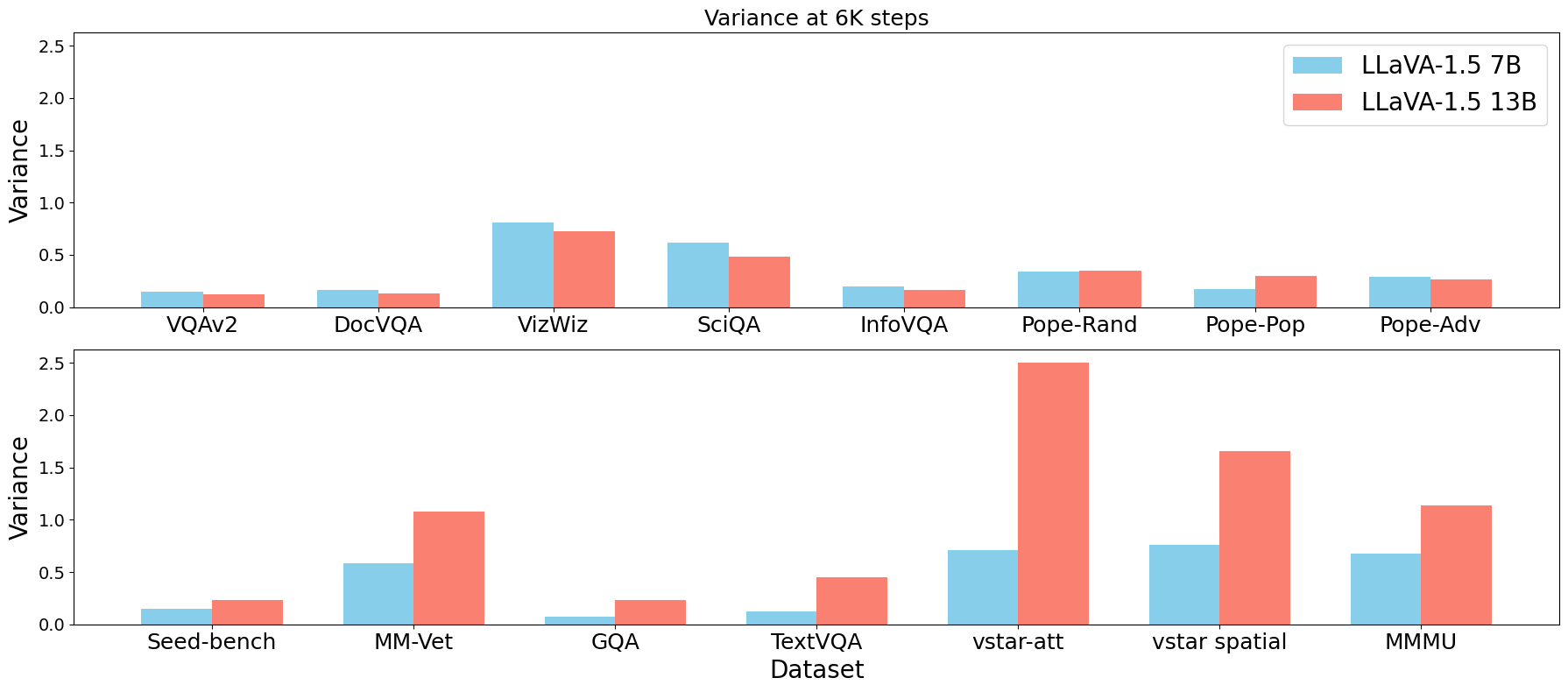}
        \caption{Variance comparison at 6K steps of finetuning.}
        \label{fig:sub2}
    \caption{Variance comparison between LLaVA-1.5-7B and 13B}
    \label{fig:largellmvar}
    \end{subfigure}
    
\end{figure}

\begin{figure}
    \centering
    \includegraphics[width=\linewidth]{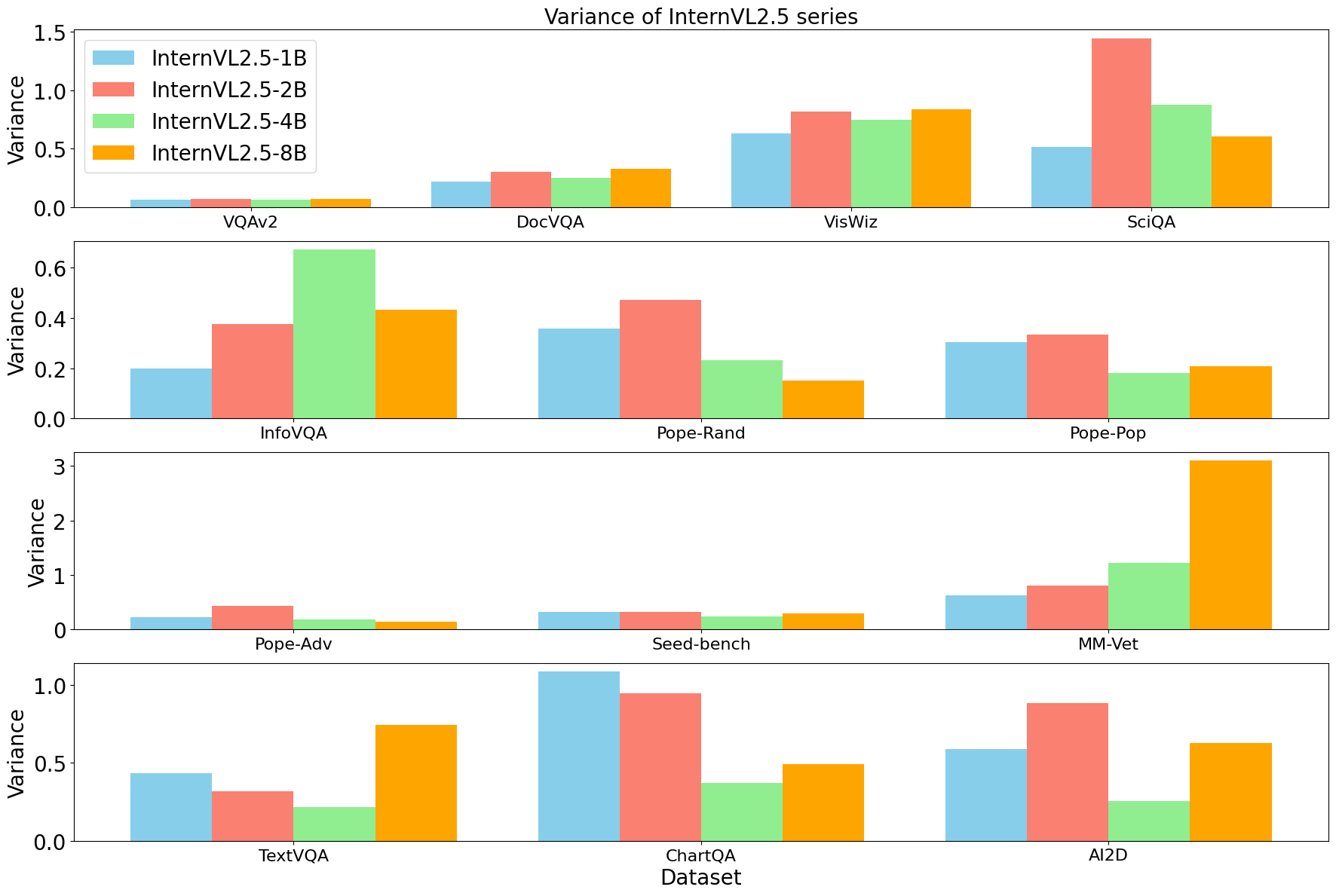}
    \caption{Variance comparison between InternVL2.5-1B, 2B, 4B and 8B}
    \label{fig:internvl_var}
\end{figure}

The size of the MLLM plays a critical role in determining its performance across various visual question answering (VQA) benchmarks. Larger MLLMs generally exhibit enhanced capabilities in reasoning, contextual understanding, and multimodal alignment, which are essential for tasks requiring complex visual and textual integration. While larger MLLMs are expected to increase the overall benchmark performances, we also check if they have increased confidence in predictions leading to decrease in variance. 
We compare the variance of LLaVA-1.5-7B and 13B\ref{fig:largellmvar}, InternVL2.5-1B, 2B, 4B and 8B \ref{fig:internvl_var}. 

However, our analysis reveals that the relationship between LLM size, performance and variance is not linear and is influenced by several factors, including the nature of the benchmark and the inherent variance in model outputs. 
\subsubsection{Performance Trends Across Benchmarks}

We systematically evaluated the impact of MLLM size on a diverse suite of VQA benchmarks, as shown in Table~\ref{tab:main-var}. The results indicate that larger models generally perform better on benchmarks requiring advanced reasoning and contextual understanding, such as MM-Vet, TextVQA, and SEED-Bench. However, the trends in variance are less consistent. Figure~\ref{fig:largellmvar} compares the variance between Vicuna-13B and Vicuna-7B, revealing that certain benchmarks including SEED-Bench, MM-Vet, GQA, TextVQA, V$^*$, and MMMU exhibit higher variance with increased model size. This suggests that while larger models often improve average performance, variance does not consistently correlate with size. In fact, variance frequently increases.

A similar trend is observed with InternVL2.5, where benchmarks such as DocVQA, VizWiz, InfographicsVQA, and MM-Vet also show increased variance with model size, as illustrated in Figure~\ref{fig:internvl_var}. This is somewhat expected for smaller benchmarks like MM-Vet and V$^*$, which are known to have inherently higher variance. Notably, both these benchmarks demonstrate a marked increase in variance as LLM size increases.

\subsection{Effect of Extended Instruction Finetuning}

\begin{figure*}[htbp]
    \centering
    \begin{tabular}{ccc}
        \includegraphics[width=0.25\textwidth]{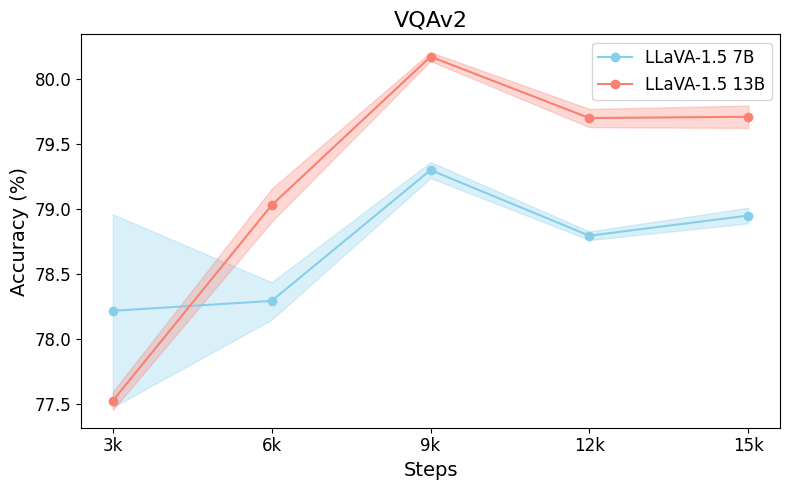} &
        \includegraphics[width=0.25\textwidth]{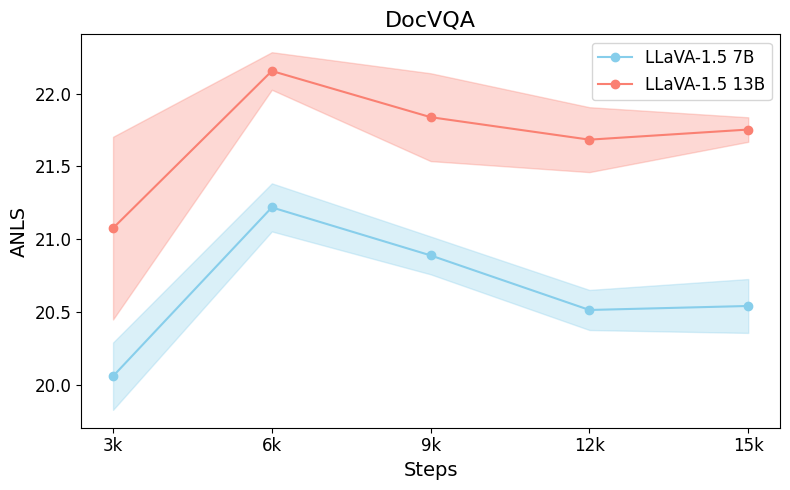} &
        \includegraphics[width=0.25\textwidth]{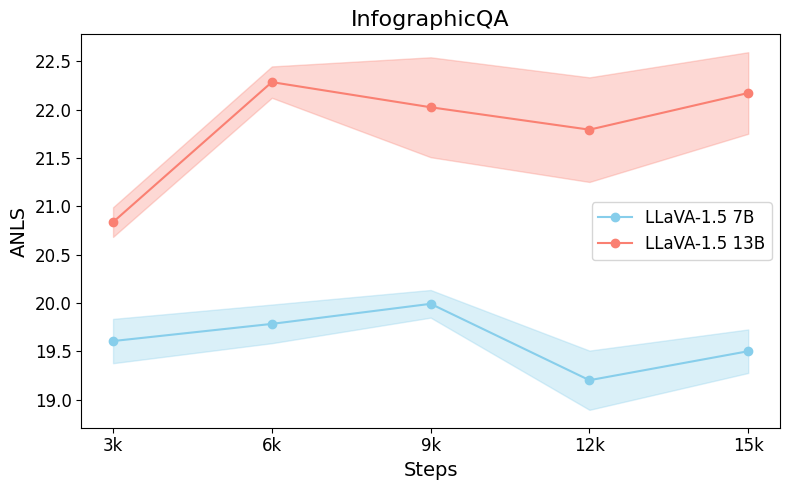} \\
        \includegraphics[width=0.25\textwidth]{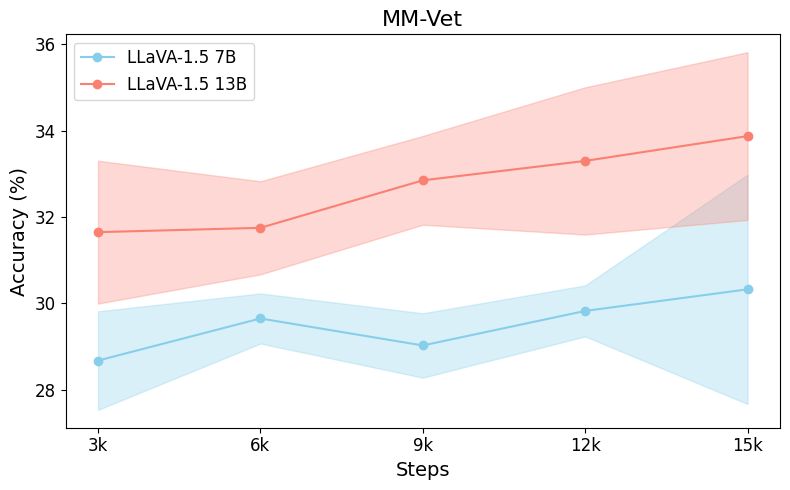} &
        \includegraphics[width=0.25\textwidth]{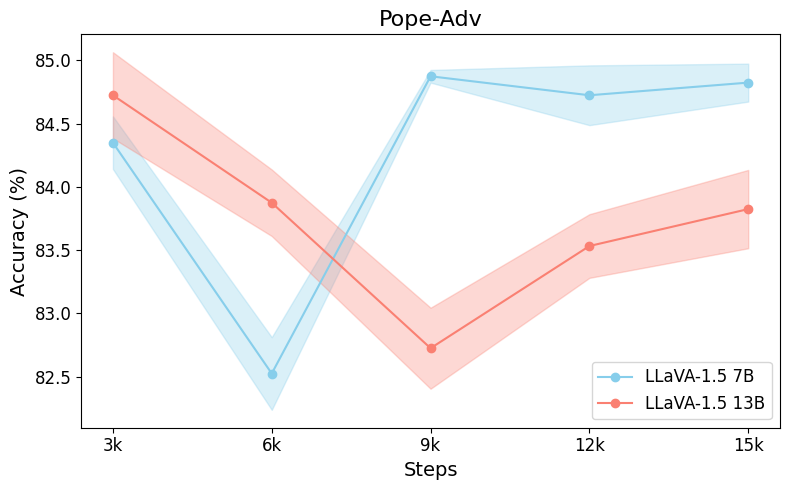} &
        \includegraphics[width=0.25\textwidth]{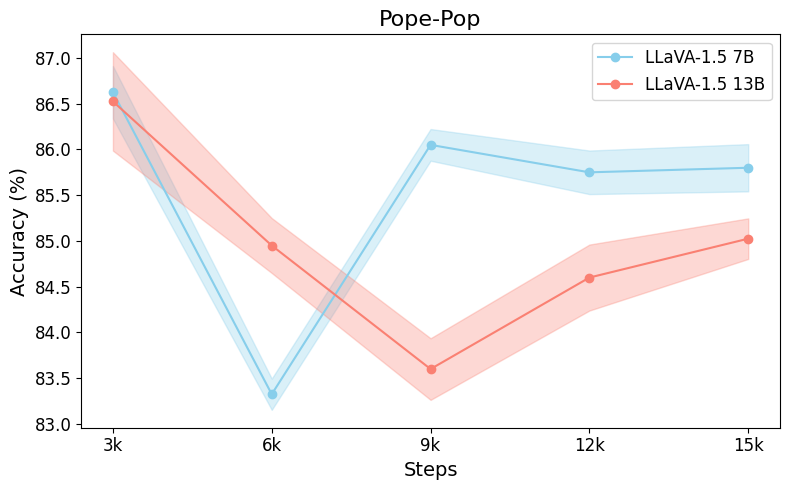} \\
        \includegraphics[width=0.25\textwidth]{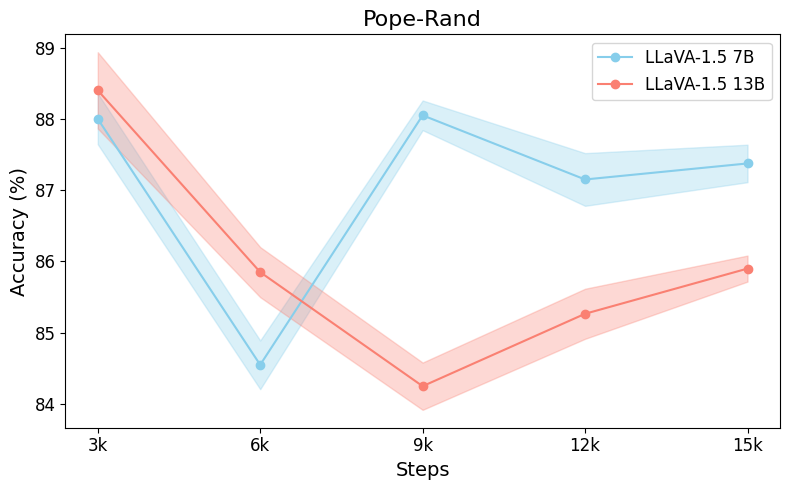} &
        \includegraphics[width=0.25\textwidth]{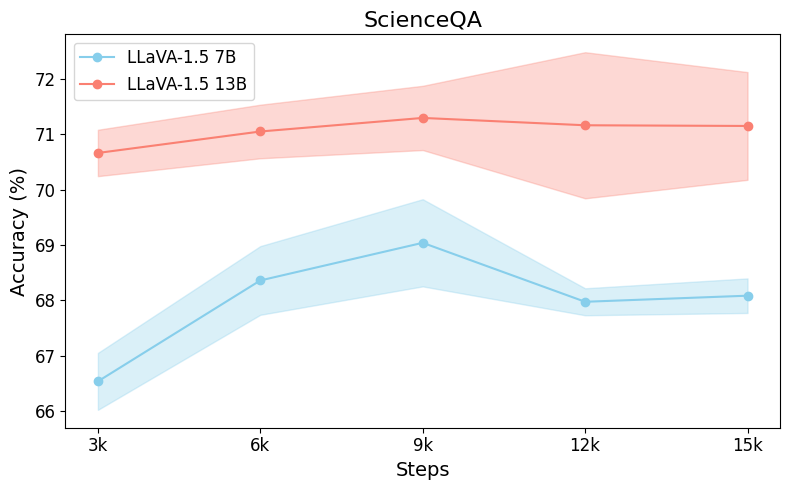} &
        \includegraphics[width=0.25\textwidth]{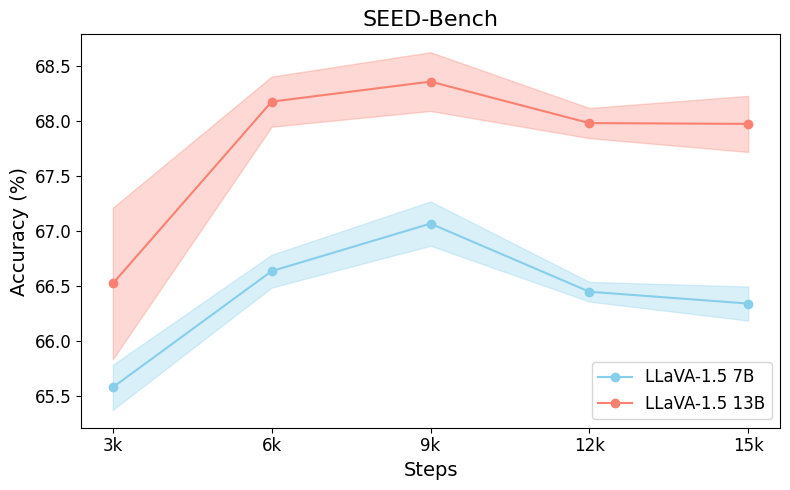} \\
        \includegraphics[width=0.25\textwidth]{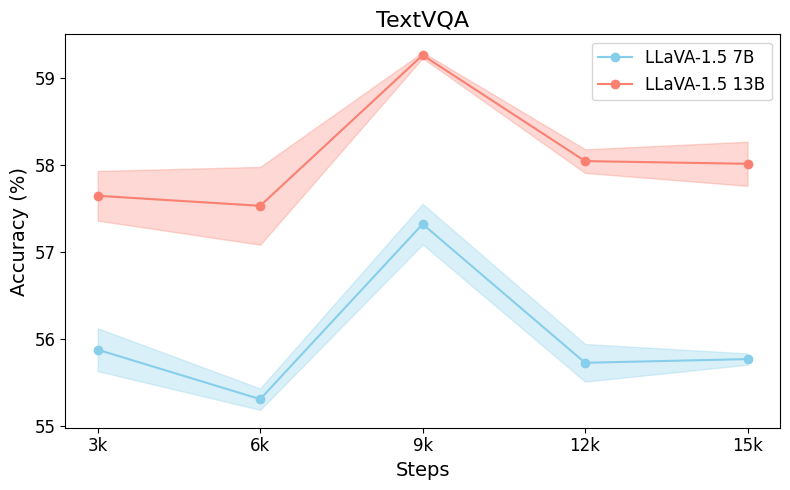} &
        \includegraphics[width=0.25\textwidth]{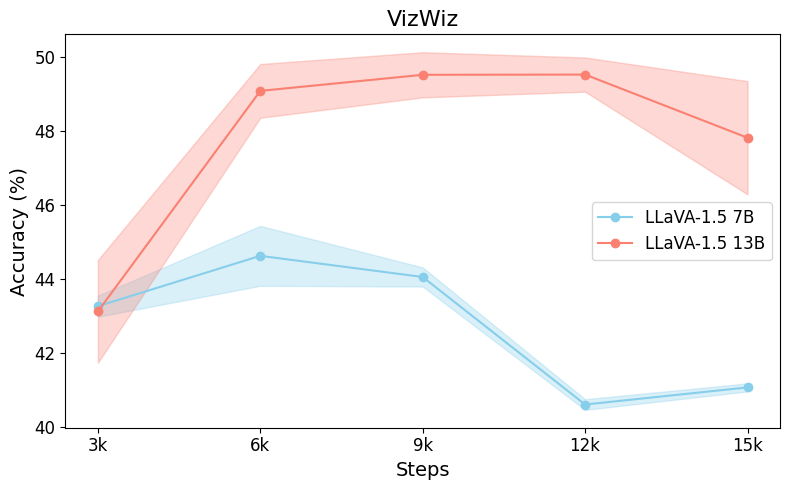} &
        \includegraphics[width=0.25\textwidth]{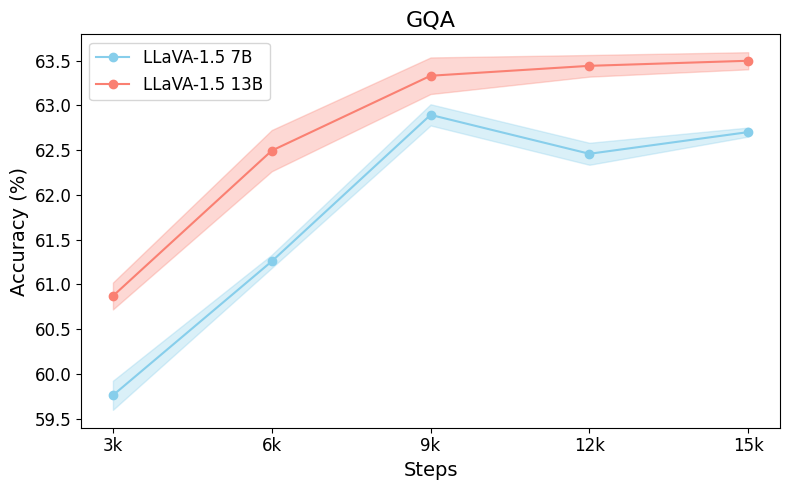} \\
        \includegraphics[width=0.25\textwidth]{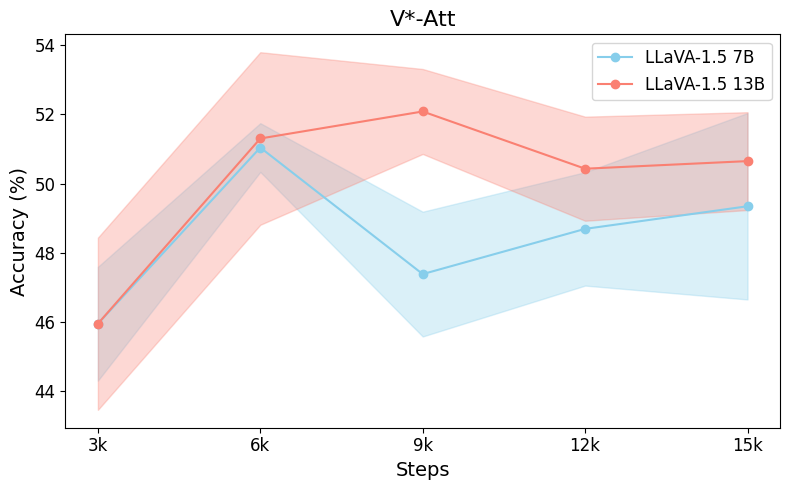} &
        \includegraphics[width=0.25\textwidth]{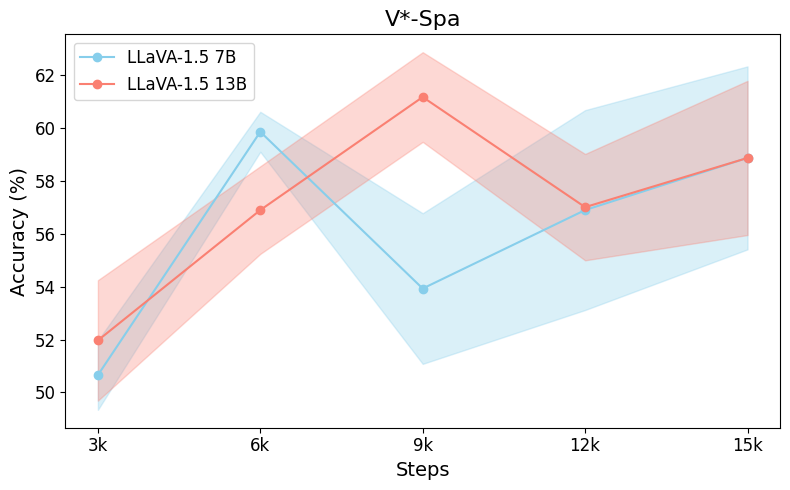} &
        \includegraphics[width=0.25\textwidth]{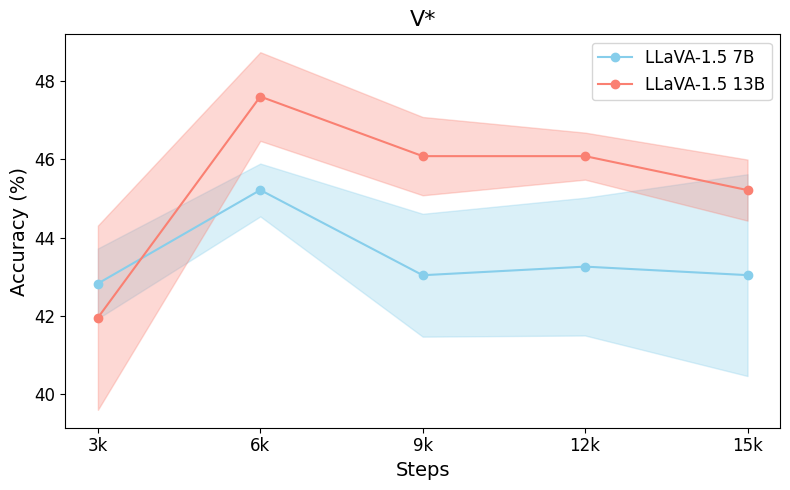} \\
    \end{tabular}
    \caption{Performance of LLaVA-1.5 7B and 13B across different finetuning steps. Each figure shows the trend observed in different benchmarks. The spread around each data point indicates the variance}
    \label{fig:performance}
\end{figure*}

\begin{figure*}[htbp]
    \centering
    \begin{tabular}{ccc}
        \includegraphics[width=0.25\textwidth]{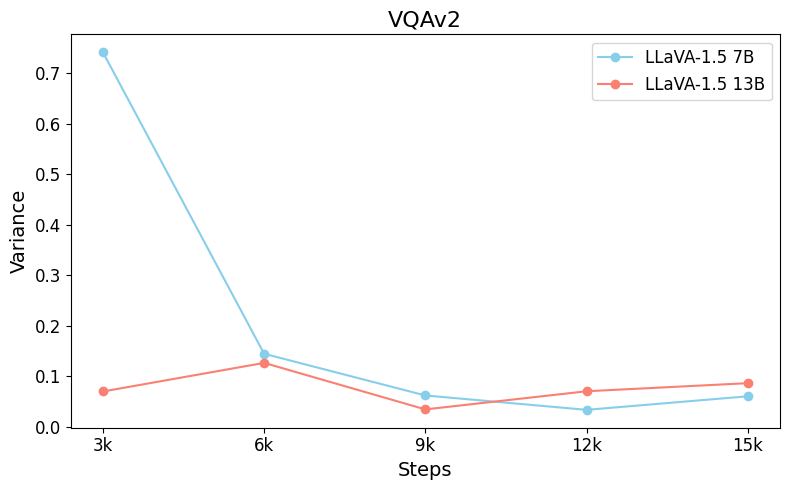} &
        \includegraphics[width=0.25\textwidth]{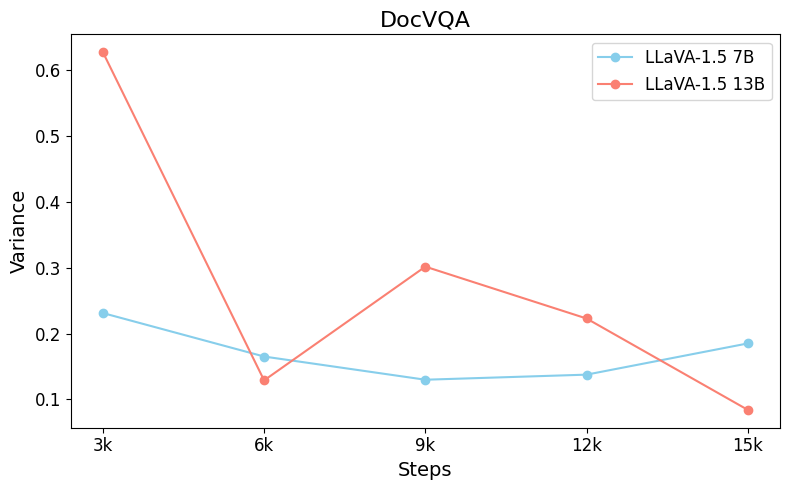} &
        \includegraphics[width=0.25\textwidth]{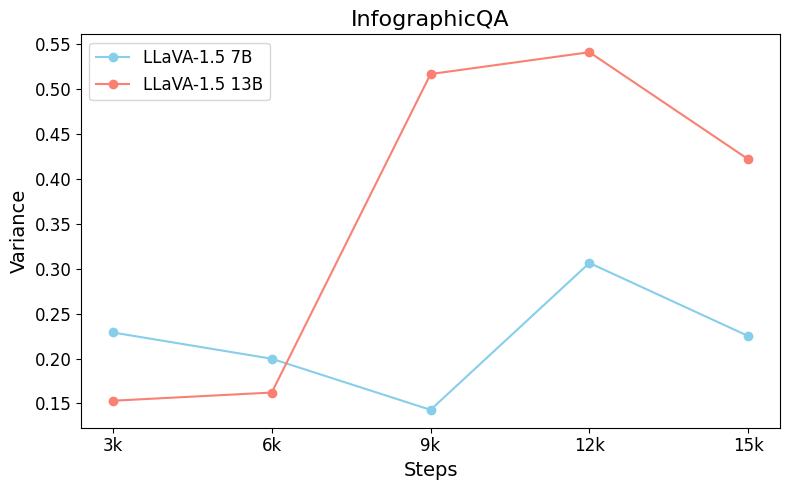} \\
        \includegraphics[width=0.25\textwidth]{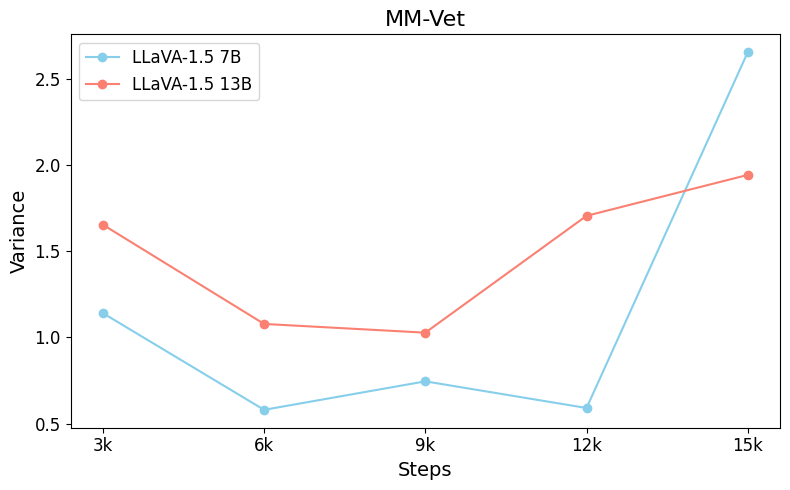} &
        \includegraphics[width=0.25\textwidth]{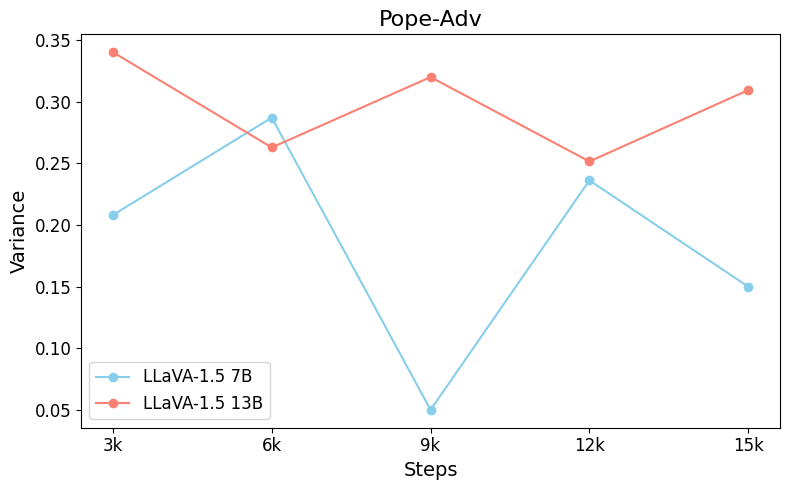} &
        \includegraphics[width=0.25\textwidth]{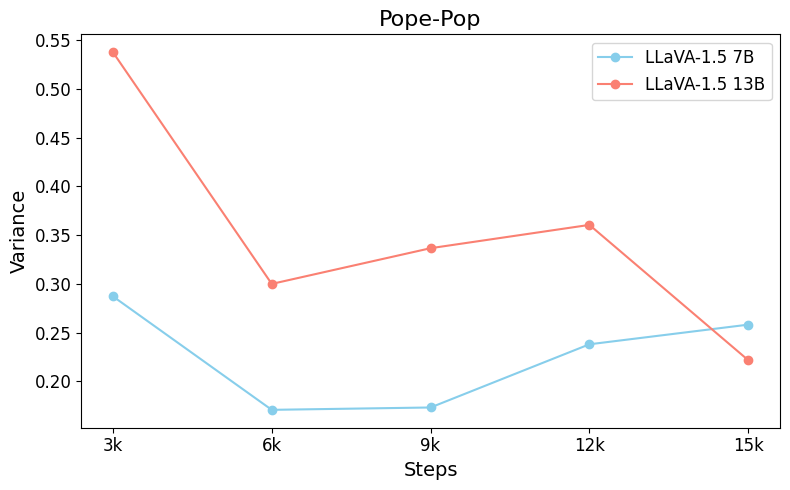} \\
        \includegraphics[width=0.25\textwidth]{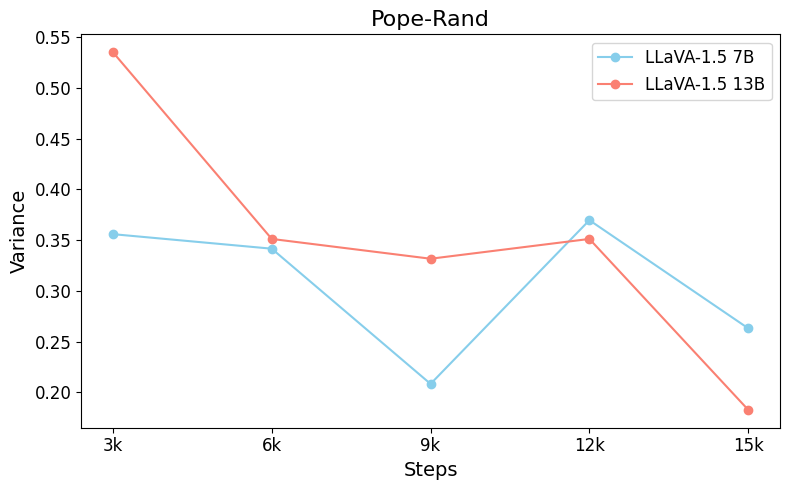} &
        \includegraphics[width=0.25\textwidth]{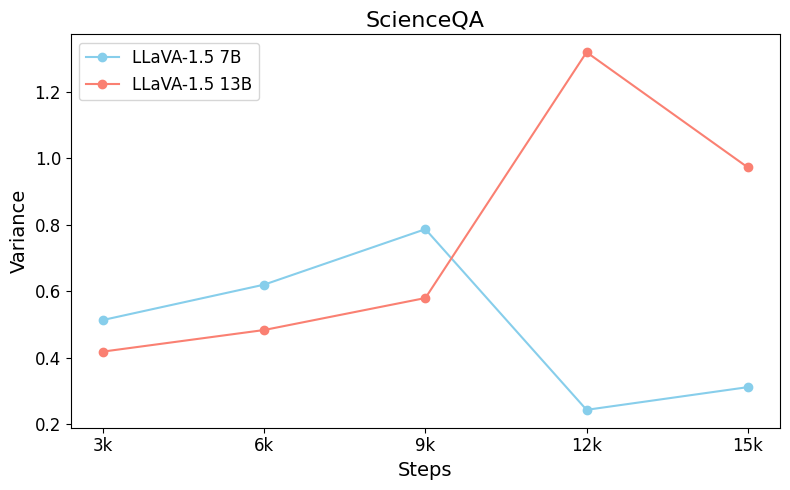} &
        \includegraphics[width=0.25\textwidth]{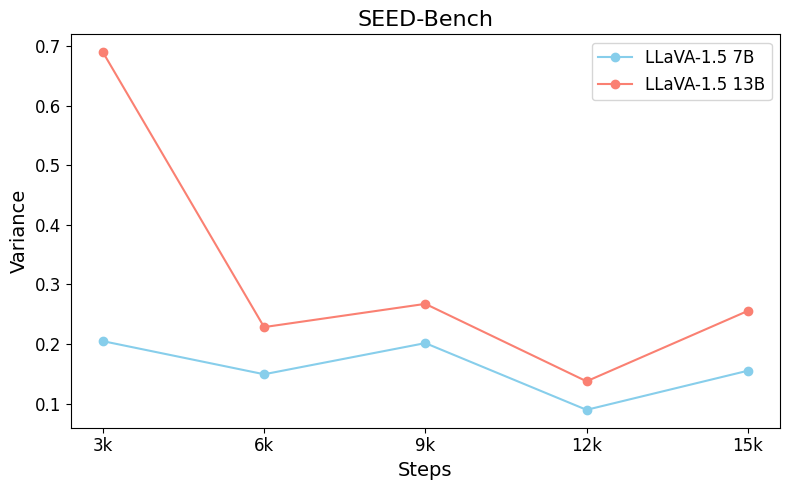} \\
        \includegraphics[width=0.25\textwidth]{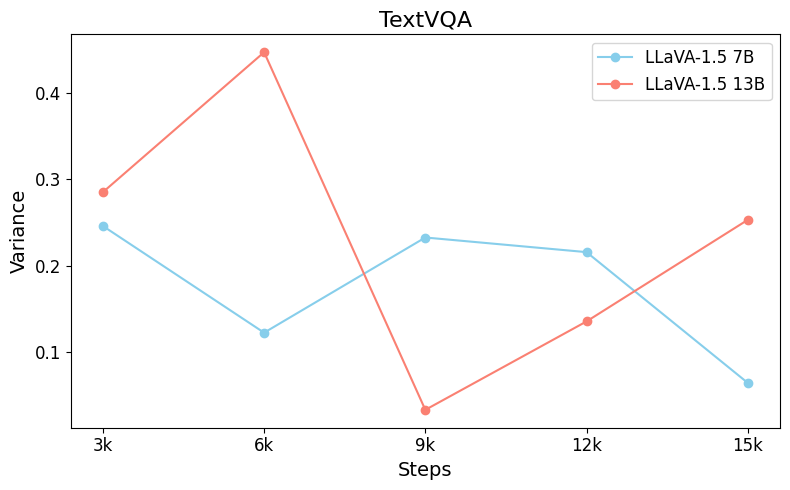} &
        \includegraphics[width=0.25\textwidth]{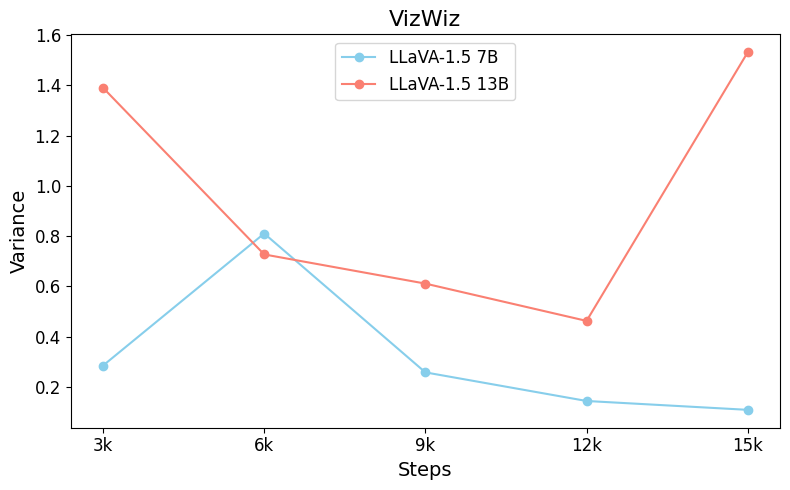} &
        \includegraphics[width=0.25\textwidth]{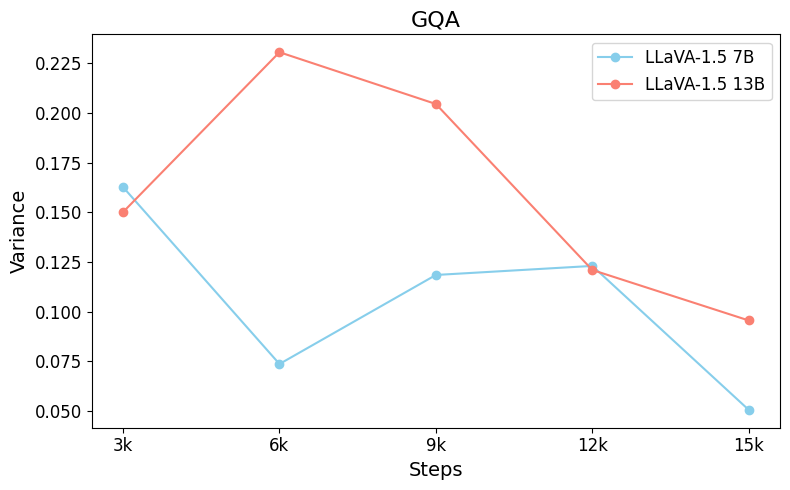} \\
    \end{tabular}
    \caption{Plots describing the trend of variance with increasing finetuning of LLaVA-1.5 7B and 13B.}
    \label{fig:variance}
\end{figure*}

The stage 2 training of MLLMs, commonly referred as instruction finetuning, is a very sensitive training stage that can influence the general performance of the models drastically. This stage always requires a fine mix of instruction following data with minimal errors. The LLM is generally unfrozen in this stage and a sufficient mix of language only data is used to retain the general language abilities of the LLM. Works like \cite{chen2024internvlscalingvisionfoundation}, \cite{liu2025nvilaefficientfrontiervisual} and \cite{bai2023qwenvlversatilevisionlanguagemodel} have sufficiently scaled the finetuning data to improve the performance of models.While these improvements in accuracy are well-documented, we sought to investigate whether extended finetuning also affects model confidence and variance. To analyse this, we extend the finetuning of LLaVA-1.5 7B and 13B to 15K steps and evaluate on intermediate checkpoints at an interval of 3K. It is to be noted that the 15K steps is essentially 3 epochs of finetuning. Certain patterns emerge as we analyse the plot in \ref{fig:performance} and \ref{fig:variance}.

\begin{itemize}
\item Comparing the performance of LLaVA-1.5 7B and 13B models in Figure~\ref{fig:performance}, we observe that the 13B variant consistently outperforms the 7B model across most benchmarks. An exception is the Pope benchmark, where the 7B model demonstrates superior performance with increasing fine-tuning steps. This trend generally aligns with the expected correlation between model size and performance.

\item Interestingly, extended instruction fine-tuning does not uniformly lead to improved downstream performance across all benchmarks. While some datasets like VQAv2, GQA, and MM-Vet exhibit a modest monotonic improvement with increased fine-tuning, others such as Pope, DocVQA, InfographicVQA, and VizWiz show stagnant or declining performance. We attribute this to the nature of the fine-tuning data, which comprises a mix of COCO, TextVQA, GQA, and Visual Genome. Consequently, benchmarks with distributions that differ significantly from the fine-tuning data, such as DocVQA, benefit less or even degrade with further training.

\item As shown in Figure~\ref{fig:variance}, the LLaVA 7B model consistently exhibits lower variance than the 13B variant in several benchmarks, including VQAv2, InfographicVQA, SEED-Bench, GQA, and V*. This trend is especially prominent at later training steps.

\item In benchmarks like ScienceQA, InfographicQA, and VizWiz, the 13B model displays a marked increase in variance over training steps, suggesting possible overfitting or heightened sensitivity to fine-tuning. In contrast, benchmarks such as VQAv2, SEED-Bench, GQA, and TextVQA that were included in the fine-tuning distribution show a steady decrease in variance as training progresses.

\item Benchmarks like MM-Vet, V*-Att, and Pope-Adv exhibit erratic variance trends across both model variants. We attribute this instability to factors such as the relatively small dataset sizes (e.g., MM-Vet and V$^*$), and the presence of noisy or inconsistent annotations, particularly in Pope, as noted by \citet{neuhaus2025repopeimpactannotationerrors}.

\item Among all benchmarks, GQA and VQAv2 consistently show the lowest variance across steps, highlighting their robustness during instruction fine-tuning.
\end{itemize}

The findings underscore the need for variance-aware practices when evaluating models trained with extended instruction finetuning.  While this technique enhances model alignment and task-specific performance, it also necessitates robust evaluation frameworks to account for residual variability.  Future research should focus on optimizing finetuning strategies to balance accuracy improvements with variance reduction.
\begin{figure}[htbp]
    \centering
    \begin{subfigure}[b]{0.23\textwidth}
        \centering
        \includegraphics[width=\textwidth]{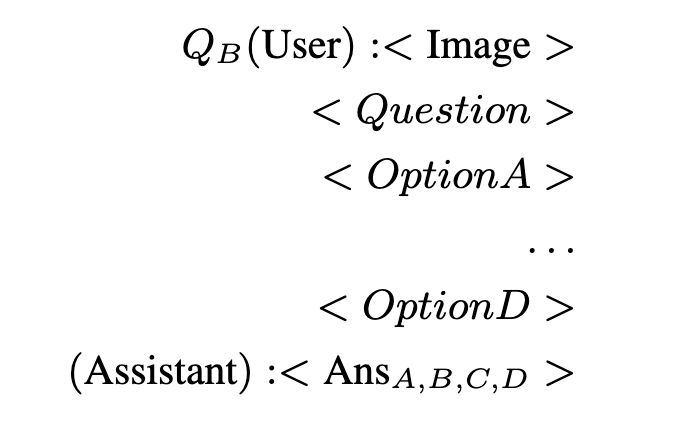}
        \caption{Normal question answering format}
        \label{fig:normal}
    \end{subfigure}
    \hfill
    \begin{subfigure}[b]{0.23\textwidth}
        \centering
        \includegraphics[width=\textwidth]{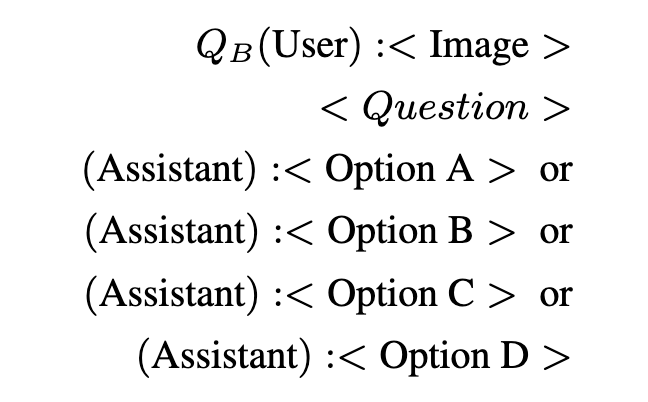}
        \caption{Cloze question answering format}
        \label{fig:cloze}
    \end{subfigure}
\end{figure}

\section{Can Cloze Evaluation Help?}

While the different aspects of model training affect the performance and variance, we observe certain amount of spuriousness in the trends in \ref{fig:performance} and \ref{fig:variance}. Hence, we go on to analyse if the variance is an effect of the evaluation method as well. \cite{madaan2024quantifyingvarianceevaluationbenchmarks} explores Cloze evaluation as an alternative to mitigate the high variance observed in MCQ based benchmarks. For any benchmark $B$ that consists of multiple choice questions the design of evaluation normally vs in cloze-style is illustrated in \ref{fig:normal} and \ref{fig:cloze}.

\begin{figure}
    \centering
    \includegraphics[width=0.9\linewidth]{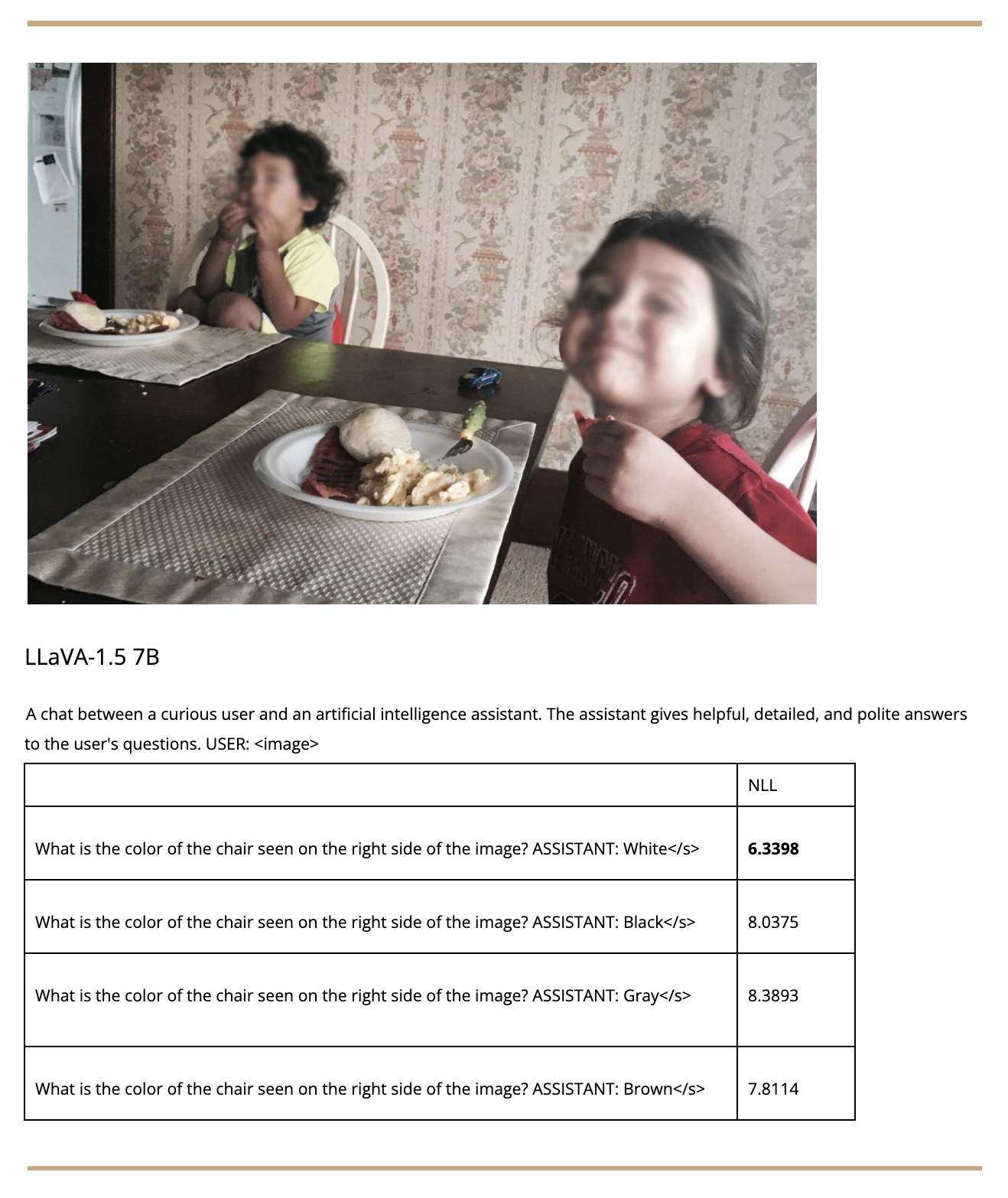}
    \caption{Sample from MMMU-Cloze evaluation.}
    \label{fig:enter-label}
\end{figure}
While the answers predicted by MLLM in $B$ are based on the direct option id's {A,B,C,D} that give the least negative log-likelihood (NLL), $B$-Cloze is evaluated by predicting the NLL of all the options directly after the question \ref{fig:enter-label}. The option with the least cross-entropy loss (ie, highest probable) is chosen as the final answer. Cloze method comes with its disadvantage as it is possible only for the multiple choice QA benchmarks. This is not applicable for other open-ended benchmarks that might require detailed answers and reasoning. Among vision benchmarks we evaluate the Cloze versions for MCQ based benchmarks: ScienceQA, Pope, SEED-Bench, MMMU and V$^*$. To perform Cloze evaluation on TextVQA, we remove the reference OCR tokens that are typically provided in the prompt and use them as potential options to compute NLL. In case of Pope, we constructed randomised options among 'Yes' and 'No'.

\begin{table}[ht]
\small
\centering
\rowcolors{2}{gray!15}{white} %
\begin{tabular}{lcc}
\toprule
\textbf{Benchmark} & $\mu(\mathcal{S},\mathbb{M})$ & $\sigma(\mathcal{S},\mathbb{M})$  \\
\midrule
ScienceQA & 69.7 & 0.62 \\
ScienceQA-C & 69.33 & 0.288 \textdownarrow\\
Pope-Rand & 88.03 & 0.568 \\
Pope-Rand-C & 87.18 \textdownarrow & 0.303 \textdownarrow\\
Pope-Pop & 86.83 & 0.9\\
Pope-Pop-C & 85.64 \textdownarrow& 0.18 \textdownarrow \\
Pope-Adv & 83.03 & 0.49 \\
Pope-Adv-C & 84.3 \textuparrow & 0.367 \textdownarrow\\
SEED  & 66.416 & 0.323\\
SEED-C  & 48.606 \textdownarrow & 0.816 \textuparrow\\
TextVQA  & 58.338 & 0.256\\
TextVQA-C  & 27.38 \textdownarrow& 0.38 \textuparrow\\
V$^*$ &  36.962 & 1.63\\
V$^*$-C & 47.958 \textuparrow & 1.26 \textdownarrow\\
MMMU & 31.7 & 0.22 \\
MMMU-C & 28.52 \textdownarrow & 0.18 \textdownarrow \\
\bottomrule
\end{tabular}
\caption{Performance across different benchmarks $B$ and their Cloze versions $B$-C.}
\label{tab:cloze}
\end{table}

\subsection{Observations}
Cloze performance (Mean and Variance) of 4 benchmarks are reported in \ref{tab:cloze}. We observe that the variance has decreased in multiple benchmarks upon using Cloze prompting technique. Except, TextVQA and SEED's variance which increased from $0.25 \rightarrow 0.38$ and $0.32 \rightarrow 0.81$, the remaining benchmarks have shown a decrease. However, rather notable effect of using Cloze evaluation is the drop in overall performance of LLaVA-1.5 7B across ScienceQA, Pope, SEED and TextVQA. This suggests that directly predicting option letters introduces inherent biases of the LLMs \cite{pezeshkpour2023largelanguagemodelssensitivity, wei-etal-2024-unveiling}, potentially leading to misleading interpretations and poorly assessed model capabilities. 

\section{Do Benchmarks Truly Test Visual Understanding?}
MLLM benchmarks are designed to check the ability of AI systems to understand images for performing tasks like question answering. While it is an enormous effort to design such benchmarks, we present a few interesting observations on these existing benchmarks. In particular , we test 1) the benchmark performances of MLLMs in blind mode where the complete training happens without the use of \texttt{<Image>} token, necessarily making it blind. This helps us uncover the possibility of MLLM benchmarks purely answerable by the world knowledge of LLMs 2) The performance of MLLMs if the vision encoder's \texttt{[CLS]} token embedding is used as the visual information. This restricts the visual input to a single token providing a global representation for the image. While prior works \cite{hu2024matryoshkaquerytransformerlarge} explore complex token reduction strategies, we find that using just the \texttt{[CLS]} token or average pooling the spatial embeddings achieves comparable performances. 

\begin{table}[ht]
\centering
\resizebox{\columnwidth}{!}{%
\begin{tabular}{c|cccc|cc}
\hline
\multicolumn{1}{c|}{\textbf{Benchmark}} & \multicolumn{4}{c|}{\textbf{LLaVA1.5-7B}} & \multicolumn{2}{c}{\textbf{LLaVA1.5-13B}} \\
\cline{2-7}
\multicolumn{1}{c|}{} & Base & CLS & Avg-pool & Blind & Base & Blind \\
\cline{2-7}
TextVQA & 58.72 & 49.04 & 46.71 & 43.33 & 62.5 & 46.43 \\

ScienceQA & 69.5 & 68.57 & 64.7 & 63.01 & 71 & 64.95 \\

Pope-R & 87.3 & 85.7 & 83.9 & 68 & 87.5 & 67.8 \\

Pope-P & 86.1 & 81.4 & 80.2 & 66.4 & 86.4 & 66.3 \\

Pope-A & 84.2 & 76.2 & 74.9 & 66.4 &85  & 66.2 \\

GQA & 62 & 53.67 & 52.23 & 40.1 & 64.7 & 40.79 \\

V$^*$-Attr & 47.95 & 29.56 & 31.3 & 26.95 & 47.6 & 26.08 \\

V$^*$-Pos & 47.95 & 48.68 & 56.57 & 51.31 & 56.9 & 52.63 \\

MM-Vet & 31.1 & 24.1 & 20.3 & 11.0 & 36.1 & 12.7 \\

SEED-Bench & 66.1 & 52.55 & 50.4 & 33.2 & 68.2 & 35.75 \\

DocVQA & 21.47 &8.6  & 7.39 & 5  & 22.15 & 5.9 \\

InfographicsVQA & 21.07 & 18.52 & 18.52 & 17.19  & 22.28 & 19.29 \\

VizWiz & 44.8 & 41.5 & 35.01 & 35.39 & 49.08 & 30.36 \\

VQAv2 & 78.5 & 68.01 & 63.88 & 45.91 & 79.03 & 46.25 \\

\hline
\end{tabular}
}
\caption{Comparison across LLaVA-1.5-7B and 13B with different configurations}
\label{tab:llava-compare}
\end{table}

\subsection{Observations}
We report the findings in \ref{tab:llava-compare}. LLaVA-1.5-7B Base consistently outperformed CLS and Avg-pool. However, these values are significant given the 576$x$ decrease in the input tokens. Using \texttt{[CLS]} embedding consistently achieves 2\% higher performance than pooling spatial CLIP embeddings. The Blind variants reveal how strongly each benchmark depends on vision tokens. In benchmarks like ScienceQA, and InfographicsVQA, the blind variant performs very close to Base model reflecting the non-necessity of the vision tokens. Other benchmarks like Pope, TextVQA and VizWiz performed significantly higher than random chance.  Only VQAv2, SEED-Bench, MM-Vet and DocVQA have a significant decline in performance. The same trend is observed in the LLaVA-1.5-13B as well. We observe TextVQA is largely reliant on the reference OCR tokens presented in the prompt along with visual tokens. These results indicate the underutilised visual tokens in LLaVA-1.5 and illustrate the variability of visual grounding across datasets.

\section{Conclusion}
In this work, we critically examine the evaluation practices of MLLMs in Visual Question Answering (VQA) benchmarks, highlighting significant performance variance caused by factors such as training seed sensitivity, non-deterministic outputs, MLLM size, and extended instruction fine-tuning. We demonstrate the limitations of point-estimate-based evaluations and the need for variance-aware methodologies. Our findings reveal that certain benchmarks like MM-Vet, V$^*$, ScienceQA and ChartQA  have large variance, larger models often introduce additional stochasticity, and extended fine-tuning does not lead to variance reduction across benchmarks, particularly for datasets with distributions differing from training data. We also find that Cloze evaluations while helpful in reducing variance, also drastically reduces performance showing the inherent option bias of MLLMs. We hope these insights motivate the development of improved evaluation protocols for open-ended tasks complementing Cloze-style methods with a focus on reducing variance and mitigating sensitivity to non-deterministic training artifacts.

\section{Acknowledgements}

I would like to thank Mausoom Sarkar, Balaji Krishnamurthy, Aradhya Neeraj Mathur and Aanisha Bhattacharyya for their valuable feedback and encouragement during the development of this work. I also gratefully acknowledge the support and resources provided by the MDSR Lab, Adobe, which were essential to conducting this research.
{
    \small
    \bibliographystyle{ieeenat_fullname}
    \bibliography{main}

\begin{thebibliography}{35}
\providecommand{\natexlab}[1]{#1}
\providecommand{\url}[1]{\texttt{#1}}
\expandafter\ifx\csname urlstyle\endcsname\relax
  \providecommand{\doi}[1]{doi: #1}\else
  \providecommand{\doi}{doi: \begingroup \urlstyle{rm}\Url}\fi

\bibitem[Bai et~al.(2023)Bai, Bai, Yang, Wang, Tan, Wang, Lin, Zhou, and Zhou]{bai2023qwenvlversatilevisionlanguagemodel}
Jinze Bai, Shuai Bai, Shusheng Yang, Shijie Wang, Sinan Tan, Peng Wang, Junyang Lin, Chang Zhou, and Jingren Zhou.
\newblock Qwen-vl: A versatile vision-language model for understanding, localization, text reading, and beyond, 2023.

\bibitem[Chen et~al.(2024{\natexlab{a}})Chen, Xu, Zhang, Liu, Liu, and Liu]{chen2024recoverablecompressionmultimodalvision}
Yi Chen, Jian Xu, Xu-Yao Zhang, Wen-Zhuo Liu, Yang-Yang Liu, and Cheng-Lin Liu.
\newblock Recoverable compression: A multimodal vision token recovery mechanism guided by text information, 2024{\natexlab{a}}.

\bibitem[Chen et~al.(2024{\natexlab{b}})Chen, Wu, Wang, Su, Chen, Xing, Zhong, Zhang, Zhu, Lu, Li, Luo, Lu, Qiao, and Dai]{chen2024internvlscalingvisionfoundation}
Zhe Chen, Jiannan Wu, Wenhai Wang, Weijie Su, Guo Chen, Sen Xing, Muyan Zhong, Qinglong Zhang, Xizhou Zhu, Lewei Lu, Bin Li, Ping Luo, Tong Lu, Yu Qiao, and Jifeng Dai.
\newblock Internvl: Scaling up vision foundation models and aligning for generic visual-linguistic tasks, 2024{\natexlab{b}}.

\bibitem[Deitke et~al.(2024)Deitke, Clark, Lee, Tripathi, Yang, Park, Salehi, Muennighoff, Lo, Soldaini, Lu, Anderson, Bransom, Ehsani, Ngo, Chen, Patel, Yatskar, Callison-Burch, Head, Hendrix, Bastani, VanderBilt, Lambert, Chou, Chheda, Sparks, Skjonsberg, Schmitz, Sarnat, Bischoff, Walsh, Newell, Wolters, Gupta, Zeng, Borchardt, Groeneveld, Nam, Lebrecht, Wittlif, Schoenick, Michel, Krishna, Weihs, Smith, Hajishirzi, Girshick, Farhadi, and Kembhavi]{deitke2024molmopixmoopenweights}
Matt Deitke, Christopher Clark, Sangho Lee, Rohun Tripathi, Yue Yang, Jae~Sung Park, Mohammadreza Salehi, Niklas Muennighoff, Kyle Lo, Luca Soldaini, Jiasen Lu, Taira Anderson, Erin Bransom, Kiana Ehsani, Huong Ngo, YenSung Chen, Ajay Patel, Mark Yatskar, Chris Callison-Burch, Andrew Head, Rose Hendrix, Favyen Bastani, Eli VanderBilt, Nathan Lambert, Yvonne Chou, Arnavi Chheda, Jenna Sparks, Sam Skjonsberg, Michael Schmitz, Aaron Sarnat, Byron Bischoff, Pete Walsh, Chris Newell, Piper Wolters, Tanmay Gupta, Kuo-Hao Zeng, Jon Borchardt, Dirk Groeneveld, Crystal Nam, Sophie Lebrecht, Caitlin Wittlif, Carissa Schoenick, Oscar Michel, Ranjay Krishna, Luca Weihs, Noah~A. Smith, Hannaneh Hajishirzi, Ross Girshick, Ali Farhadi, and Aniruddha Kembhavi.
\newblock Molmo and pixmo: Open weights and open data for state-of-the-art vision-language models, 2024.

\bibitem[Goyal et~al.(2017)Goyal, Khot, Summers-Stay, Batra, and Parikh]{goyal2017makingvvqamatter}
Yash Goyal, Tejas Khot, Douglas Summers-Stay, Dhruv Batra, and Devi Parikh.
\newblock Making the v in vqa matter: Elevating the role of image understanding in visual question answering, 2017.

\bibitem[Gurari et~al.(2018)Gurari, Li, Stangl, Guo, Lin, Grauman, Luo, and Bigham]{gurari2018vizwizgrandchallengeanswering}
Danna Gurari, Qing Li, Abigale~J. Stangl, Anhong Guo, Chi Lin, Kristen Grauman, Jiebo Luo, and Jeffrey~P. Bigham.
\newblock Vizwiz grand challenge: Answering visual questions from blind people, 2018.

\bibitem[Hong et~al.(2024)Hong, Wang, Ding, Yu, Lv, Wang, Cheng, Huang, Ji, Xue, Zhao, Yang, Gu, Zhang, Feng, Yin, Wang, Qi, Song, Zhang, Liu, Xu, Li, Dong, and Tang]{hong2024cogvlm2visuallanguagemodels}
Wenyi Hong, Weihan Wang, Ming Ding, Wenmeng Yu, Qingsong Lv, Yan Wang, Yean Cheng, Shiyu Huang, Junhui Ji, Zhao Xue, Lei Zhao, Zhuoyi Yang, Xiaotao Gu, Xiaohan Zhang, Guanyu Feng, Da Yin, Zihan Wang, Ji Qi, Xixuan Song, Peng Zhang, Debing Liu, Bin Xu, Juanzi Li, Yuxiao Dong, and Jie Tang.
\newblock Cogvlm2: Visual language models for image and video understanding, 2024.

\bibitem[Hu et~al.(2024)Hu, Dou, Li, Kamath, Peng, and Chang]{hu2024matryoshkaquerytransformerlarge}
Wenbo Hu, Zi-Yi Dou, Liunian~Harold Li, Amita Kamath, Nanyun Peng, and Kai-Wei Chang.
\newblock Matryoshka query transformer for large vision-language models, 2024.

\bibitem[Hudson and Manning(2019)]{hudson2019gqanewdatasetrealworld}
Drew~A. Hudson and Christopher~D. Manning.
\newblock Gqa: A new dataset for real-world visual reasoning and compositional question answering, 2019.

\bibitem[Kembhavi et~al.(2016)Kembhavi, Salvato, Kolve, Seo, Hajishirzi, and Farhadi]{Kembhavi2016ADI}
Aniruddha Kembhavi, Michael Salvato, Eric Kolve, Minjoon Seo, Hannaneh Hajishirzi, and Ali Farhadi.
\newblock A diagram is worth a dozen images.
\newblock \emph{ArXiv}, abs/1603.07396, 2016.

\bibitem[Li et~al.(2023{\natexlab{a}})Li, Wang, Wang, Ge, Ge, and Shan]{li2023seedbenchbenchmarkingmultimodalllms}
Bohao Li, Rui Wang, Guangzhi Wang, Yuying Ge, Yixiao Ge, and Ying Shan.
\newblock Seed-bench: Benchmarking multimodal llms with generative comprehension, 2023{\natexlab{a}}.

\bibitem[Li et~al.(2024)Li, Zhang, Guo, Zhang, Li, Zhang, Zhang, Zhang, Li, Liu, and Li]{li2024llavaonevisioneasyvisualtask}
Bo Li, Yuanhan Zhang, Dong Guo, Renrui Zhang, Feng Li, Hao Zhang, Kaichen Zhang, Peiyuan Zhang, Yanwei Li, Ziwei Liu, and Chunyuan Li.
\newblock Llava-onevision: Easy visual task transfer, 2024.

\bibitem[Li et~al.(2023{\natexlab{b}})Li, Du, Zhou, Wang, Zhao, and Wen]{li2023evaluatingobjecthallucinationlarge}
Yifan Li, Yifan Du, Kun Zhou, Jinpeng Wang, Wayne~Xin Zhao, and Ji-Rong Wen.
\newblock Evaluating object hallucination in large vision-language models, 2023{\natexlab{b}}.

\bibitem[Lin et~al.(2024)Lin, Yin, Ping, Lu, Molchanov, Tao, Mao, Kautz, Shoeybi, and Han]{lin2024vilapretrainingvisuallanguage}
Ji Lin, Hongxu Yin, Wei Ping, Yao Lu, Pavlo Molchanov, Andrew Tao, Huizi Mao, Jan Kautz, Mohammad Shoeybi, and Song Han.
\newblock Vila: On pre-training for visual language models, 2024.

\bibitem[Liu et~al.(2024)Liu, Li, Li, and Lee]{liu2024improvedbaselinesvisualinstruction}
Haotian Liu, Chunyuan Li, Yuheng Li, and Yong~Jae Lee.
\newblock Improved baselines with visual instruction tuning, 2024.

\bibitem[Liu et~al.(2025)Liu, Zhu, Shi, Zhang, Lou, Yang, Xi, Cao, Gu, Li, Li, Fang, Chen, Hsieh, Huang, Cheng, Nath, Hu, Liu, Krishna, Xu, Wang, Molchanov, Kautz, Yin, Han, and Lu]{liu2025nvilaefficientfrontiervisual}
Zhijian Liu, Ligeng Zhu, Baifeng Shi, Zhuoyang Zhang, Yuming Lou, Shang Yang, Haocheng Xi, Shiyi Cao, Yuxian Gu, Dacheng Li, Xiuyu Li, Yunhao Fang, Yukang Chen, Cheng-Yu Hsieh, De-An Huang, An-Chieh Cheng, Vishwesh Nath, Jinyi Hu, Sifei Liu, Ranjay Krishna, Daguang Xu, Xiaolong Wang, Pavlo Molchanov, Jan Kautz, Hongxu Yin, Song Han, and Yao Lu.
\newblock Nvila: Efficient frontier visual language models, 2025.

\bibitem[Lu et~al.(2024)Lu, Liu, Zhang, Wang, Dong, Liu, Sun, Ren, Li, Yang, Sun, Deng, Xu, Xie, and Ruan]{lu2024deepseekvlrealworldvisionlanguageunderstanding}
Haoyu Lu, Wen Liu, Bo Zhang, Bingxuan Wang, Kai Dong, Bo Liu, Jingxiang Sun, Tongzheng Ren, Zhuoshu Li, Hao Yang, Yaofeng Sun, Chengqi Deng, Hanwei Xu, Zhenda Xie, and Chong Ruan.
\newblock Deepseek-vl: Towards real-world vision-language understanding, 2024.

\bibitem[Lu et~al.(2022)Lu, Mishra, Xia, Qiu, Chang, Zhu, Tafjord, Clark, and Kalyan]{lu2022learn}
Pan Lu, Swaroop Mishra, Tony Xia, Liang Qiu, Kai-Wei Chang, Song-Chun Zhu, Oyvind Tafjord, Peter Clark, and Ashwin Kalyan.
\newblock Learn to explain: Multimodal reasoning via thought chains for science question answering.
\newblock In \emph{The 36th Conference on Neural Information Processing Systems (NeurIPS)}, 2022.

\bibitem[Madaan et~al.(2024)Madaan, Singh, Schaeffer, Poulton, Koyejo, Stenetorp, Narang, and Hupkes]{madaan2024quantifyingvarianceevaluationbenchmarks}
Lovish Madaan, Aaditya~K. Singh, Rylan Schaeffer, Andrew Poulton, Sanmi Koyejo, Pontus Stenetorp, Sharan Narang, and Dieuwke Hupkes.
\newblock Quantifying variance in evaluation benchmarks, 2024.

\bibitem[Marafioti et~al.(2025)Marafioti, Zohar, Farré, Noyan, Bakouch, Cuenca, Zakka, Allal, Lozhkov, Tazi, Srivastav, Lochner, Larcher, Morlon, Tunstall, von Werra, and Wolf]{marafioti2025smolvlmredefiningsmallefficient}
Andrés Marafioti, Orr Zohar, Miquel Farré, Merve Noyan, Elie Bakouch, Pedro Cuenca, Cyril Zakka, Loubna~Ben Allal, Anton Lozhkov, Nouamane Tazi, Vaibhav Srivastav, Joshua Lochner, Hugo Larcher, Mathieu Morlon, Lewis Tunstall, Leandro von Werra, and Thomas Wolf.
\newblock Smolvlm: Redefining small and efficient multimodal models, 2025.

\bibitem[Masry et~al.(2022)Masry, Long, Tan, Joty, and Hoque]{masry2022chartqabenchmarkquestionanswering}
Ahmed Masry, Do~Xuan Long, Jia~Qing Tan, Shafiq Joty, and Enamul Hoque.
\newblock Chartqa: A benchmark for question answering about charts with visual and logical reasoning, 2022.

\bibitem[Mathew et~al.(2021{\natexlab{a}})Mathew, Bagal, Tito, Karatzas, Valveny, and Jawahar]{mathew2021infographicvqa}
Minesh Mathew, Viraj Bagal, Rubèn~Pérez Tito, Dimosthenis Karatzas, Ernest Valveny, and C.~V Jawahar.
\newblock Infographicvqa, 2021{\natexlab{a}}.

\bibitem[Mathew et~al.(2021{\natexlab{b}})Mathew, Karatzas, and Jawahar]{mathew2021docvqadatasetvqadocument}
Minesh Mathew, Dimosthenis Karatzas, and C.~V. Jawahar.
\newblock Docvqa: A dataset for vqa on document images, 2021{\natexlab{b}}.

\bibitem[Neuhaus and Hein(2025)]{neuhaus2025repopeimpactannotationerrors}
Yannic Neuhaus and Matthias Hein.
\newblock Repope: Impact of annotation errors on the pope benchmark, 2025.

\bibitem[Pezeshkpour and Hruschka(2023)]{pezeshkpour2023largelanguagemodelssensitivity}
Pouya Pezeshkpour and Estevam Hruschka.
\newblock Large language models sensitivity to the order of options in multiple-choice questions, 2023.

\bibitem[Shi et~al.(2024)Shi, Wu, Mao, Wang, and Darrell]{shi2024needlargervisionmodels}
Baifeng Shi, Ziyang Wu, Maolin Mao, Xin Wang, and Trevor Darrell.
\newblock When do we not need larger vision models?, 2024.

\bibitem[Singh et~al.(2019)Singh, Natarajan, Shah, Jiang, Chen, Batra, Parikh, and Rohrbach]{singh2019vqamodelsread}
Amanpreet Singh, Vivek Natarajan, Meet Shah, Yu Jiang, Xinlei Chen, Dhruv Batra, Devi Parikh, and Marcus Rohrbach.
\newblock Towards vqa models that can read, 2019.

\bibitem[SR et~al.(2025)SR, Mathur, Menta, Jain, and Sarkar]{sr2025hirelightweighthighresolutionimage}
Nikitha SR, Aradhya~Neeraj Mathur, Tarun~Ram Menta, Rishabh Jain, and Mausoom Sarkar.
\newblock Hire: Lightweight high-resolution image feature enrichment for multimodal llms, 2025.

\bibitem[Wei et~al.(2024)Wei, Wu, Huang, and Chen]{wei-etal-2024-unveiling}
Sheng-Lun Wei, Cheng-Kuang Wu, Hen-Hsen Huang, and Hsin-Hsi Chen.
\newblock Unveiling selection biases: Exploring order and token sensitivity in large language models.
\newblock In \emph{Findings of the Association for Computational Linguistics: ACL 2024}, pages 5598--5621, Bangkok, Thailand, 2024. Association for Computational Linguistics.

\bibitem[Wu and Xie(2023)]{wu2023vguidedvisualsearch}
Penghao Wu and Saining Xie.
\newblock V*: Guided visual search as a core mechanism in multimodal llms, 2023.

\bibitem[Yao et~al.(2024)Yao, Yu, Zhang, Wang, Cui, Zhu, Cai, Li, Zhao, He, Chen, Zhou, Zou, Zhang, Hu, Zheng, Zhou, Cai, Han, Zeng, Li, Liu, and Sun]{yao2024minicpmvgpt4vlevelmllm}
Yuan Yao, Tianyu Yu, Ao Zhang, Chongyi Wang, Junbo Cui, Hongji Zhu, Tianchi Cai, Haoyu Li, Weilin Zhao, Zhihui He, Qianyu Chen, Huarong Zhou, Zhensheng Zou, Haoye Zhang, Shengding Hu, Zhi Zheng, Jie Zhou, Jie Cai, Xu Han, Guoyang Zeng, Dahai Li, Zhiyuan Liu, and Maosong Sun.
\newblock Minicpm-v: A gpt-4v level mllm on your phone, 2024.

\bibitem[Ye et~al.(2024)Ye, Xu, Liu, Hu, Yan, Qian, Zhang, Huang, and Zhou]{ye2024mplugowl3longimagesequenceunderstanding}
Jiabo Ye, Haiyang Xu, Haowei Liu, Anwen Hu, Ming Yan, Qi Qian, Ji Zhang, Fei Huang, and Jingren Zhou.
\newblock mplug-owl3: Towards long image-sequence understanding in multi-modal large language models, 2024.

\bibitem[Ying et~al.(2024)Ying, Meng, Wang, Li, Lin, Yang, Zhang, Zhang, Lin, Liu, Lei, Lu, Chen, Xu, Zhang, Zhang, Gao, Wang, Qiao, Luo, Zhang, and Shao]{ying2024mmtbenchcomprehensivemultimodalbenchmark}
Kaining Ying, Fanqing Meng, Jin Wang, Zhiqian Li, Han Lin, Yue Yang, Hao Zhang, Wenbo Zhang, Yuqi Lin, Shuo Liu, Jiayi Lei, Quanfeng Lu, Runjian Chen, Peng Xu, Renrui Zhang, Haozhe Zhang, Peng Gao, Yali Wang, Yu Qiao, Ping Luo, Kaipeng Zhang, and Wenqi Shao.
\newblock Mmt-bench: A comprehensive multimodal benchmark for evaluating large vision-language models towards multitask agi, 2024.

\bibitem[Yu et~al.(2024)Yu, Yang, Li, Wang, Lin, Liu, Wang, and Wang]{yu2024mmvetevaluatinglargemultimodal}
Weihao Yu, Zhengyuan Yang, Linjie Li, Jianfeng Wang, Kevin Lin, Zicheng Liu, Xinchao Wang, and Lijuan Wang.
\newblock Mm-vet: Evaluating large multimodal models for integrated capabilities, 2024.

\bibitem[Yue et~al.(2024)Yue, Ni, Zhang, Zheng, Liu, Zhang, Stevens, Jiang, Ren, Sun, Wei, Yu, Yuan, Sun, Yin, Zheng, Yang, Liu, Huang, Sun, Su, and Chen]{yue2024mmmumassivemultidisciplinemultimodal}
Xiang Yue, Yuansheng Ni, Kai Zhang, Tianyu Zheng, Ruoqi Liu, Ge Zhang, Samuel Stevens, Dongfu Jiang, Weiming Ren, Yuxuan Sun, Cong Wei, Botao Yu, Ruibin Yuan, Renliang Sun, Ming Yin, Boyuan Zheng, Zhenzhu Yang, Yibo Liu, Wenhao Huang, Huan Sun, Yu Su, and Wenhu Chen.
\newblock Mmmu: A massive multi-discipline multimodal understanding and reasoning benchmark for expert agi, 2024.

\end{thebibliography}
}

\end{document}